%% file: ms.tex
\documentclass[10pt,twocolumn,letterpaper]{article}

\usepackage{cvpr}
\usepackage{times}
\usepackage{epsfig}
\usepackage{graphicx}
\usepackage{amsmath}
\usepackage{amssymb}
\usepackage{textcomp}
\usepackage{subcaption}
\usepackage{xargs}
\usepackage{threeparttable}
\usepackage{makecell}
\usepackage{booktabs}
\usepackage{multirow}
\usepackage{arydshln}
\usepackage[pagebackref=true,breaklinks=true,colorlinks,bookmarks=false]{hyperref} %letterpaper=true

\usepackage[capitalise]{cleveref}

\crefname{table}{table}{tables}
\Crefname{table}{Table}{Tables}
\crefname{figure}{figure}{figures}
\Crefname{figure}{Figure}{Figures}
\crefname{section}{section}{sections}
\Crefname{section}{Section}{Sections}
\crefname{subsection}{subsection}{subsections}
\Crefname{subsection}{Subsection}{Subsections}
\crefname{chapter}{chapter}{chapters}
\Crefname{chapter}{Chapter}{Chapters}
\crefname{appendix}{appendix}{appendices}
\Crefname{appendix}{Appendix}{Appendices}

\cvprfinalcopy % *** Uncomment this line for the final submission

\def\cvprPaperID{5} % *** Enter the CVPR Paper ID here

\ifcvprfinal\fi
\begin{document}

\title{Detecting CNN-Generated Facial Images in Real-World Scenarios}
\author{Nils Hulzebosch$^{1,2}$ \quad Sarah Ibrahimi$^{1,2}$ \quad Marcel Worring$^1$ \\
$^1$University of Amsterdam \qquad $^2$Dutch National Police \\}
\maketitle
\thispagestyle{empty}

\input{sections/abstract.tex}
\input{sections/introduction.tex}
\input{sections/related_work.tex}
\input{sections/methods.tex}
\input{sections/onlinesurvey.tex}
\input{sections/results.tex}
\input{sections/conclusion.tex}
\clearpage

{\small
\bibliographystyle{ieee_fullname}
\bibliography{ms}
}

\end{document}

\title{Detecting CNN-Generated Facial Images in Real-World Scenarios\\
{\Large \fontseries{m} \fontshape{n} \textsc{Supplementary Material}}}
%\title{SUPPLEMENTARY MATERIAL}
\author{Nils Hulzebosch$^{1,2}$ \quad Sarah Ibrahimi$^{1,2}$ \quad Marcel Worring$^1$ \\
$^1$University of Amsterdam \qquad $^2$Dutch National Police \\}
\maketitle
\thispagestyle{empty}

\input{sections/appendix.tex}

%% file: sections/abstract.tex
%%%%%%%%% ABSTRACT
\begin{abstract}
   Artificial, CNN-generated images are now of such high quality that humans have trouble distinguishing them from real images. Several algorithmic detection methods have been proposed, but these appear to generalize poorly to data from unknown sources, making them infeasible for real-world scenarios.
   In this work, we present a framework for evaluating detection methods under real-world conditions, consisting of cross-model, cross-data, and post-processing evaluation, and we evaluate state-of-the-art detection methods using the proposed framework. Furthermore, we examine the usefulness of commonly used image pre-processing methods. Lastly, we evaluate human performance on detecting CNN-generated images, along with factors that influence this performance, by conducting an online survey. Our results suggest that CNN-based detection methods are not yet robust enough to be used in real-world scenarios.
\end{abstract}

%% file: sections/introduction.tex
\section{Introduction}
\label{sec:introduction}
Recently, state-of-the-art CNN-based generative models have radically improved the visual quality of generated images \cite{karras2019style, karras2019analyzing}. Combined with an increasing ease of using such models by
non-experts through user friendly applications (e.g. \cite{cole2018blog,cole2019blog,porter2019blog}), there is sufficient reason to be cautious about its use by people with harmful intents. The malicious use of technologies employing generative models has been demonstrated with DeepFakes in the form of (revenge) pornography, where faces of women are mapped to pornographic videos \cite{cole2018blog}, and with DeepNude by undressing women \cite{cole2019blog}. The potential of DeepFakes for political purposes has also been demonstrated in \cite{cole2019blog_political,gilmer2019blog,parkin2019blog}, and has the capability to become a significant problem in terms of fake news and propaganda. 
Current state-of-the-art generative models \cite{karras2019style,karras2019analyzing} go one step further and are capable of creating fully-generated realistic images of human faces. The development of image generation techniques will likely have ethical, moral, and legal consequences.
\begin{figure}[!t]
    \centering
    \captionsetup{width=\linewidth}
    \captionsetup{font=small}
    \begin{subfigure}[b]{0.24\linewidth}
        \includegraphics[width=\linewidth]{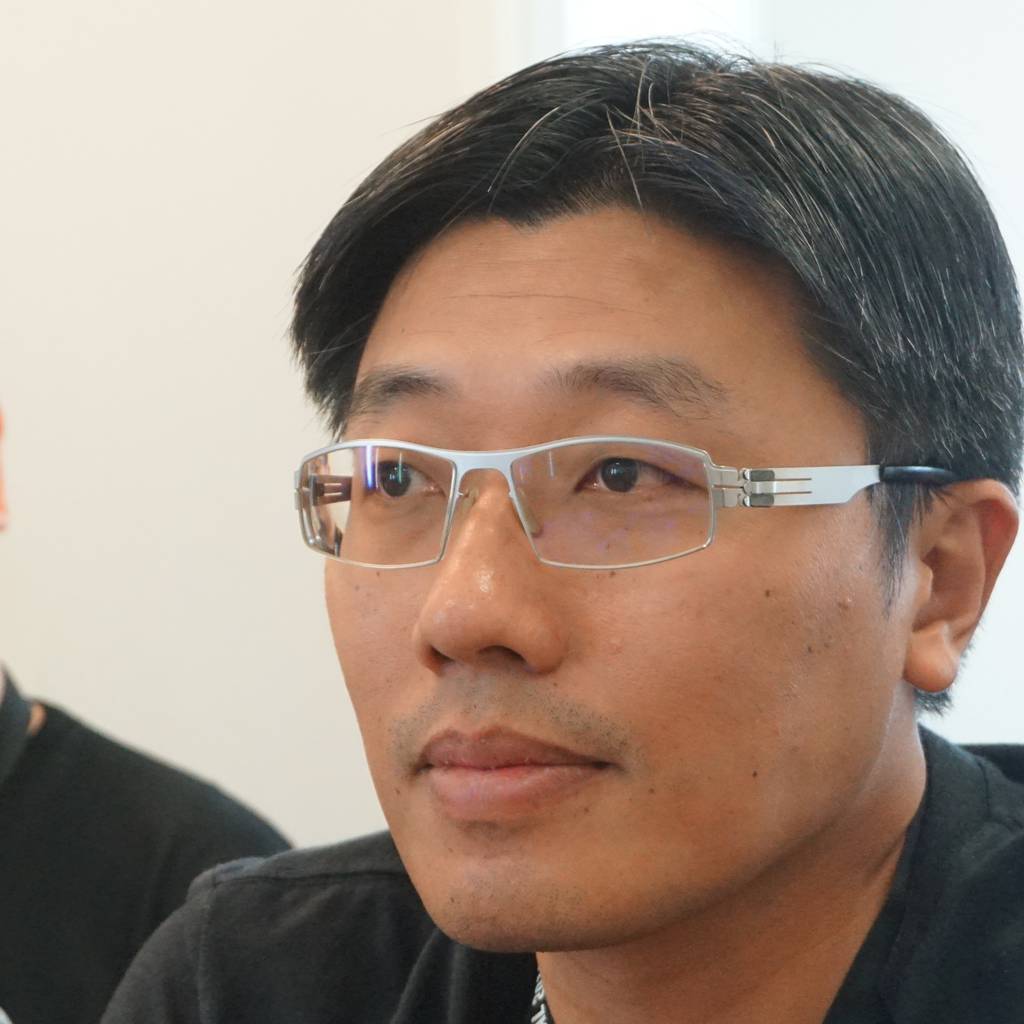}
    \end{subfigure}
    \begin{subfigure}[b]{0.24\linewidth}
        \includegraphics[width=\linewidth]{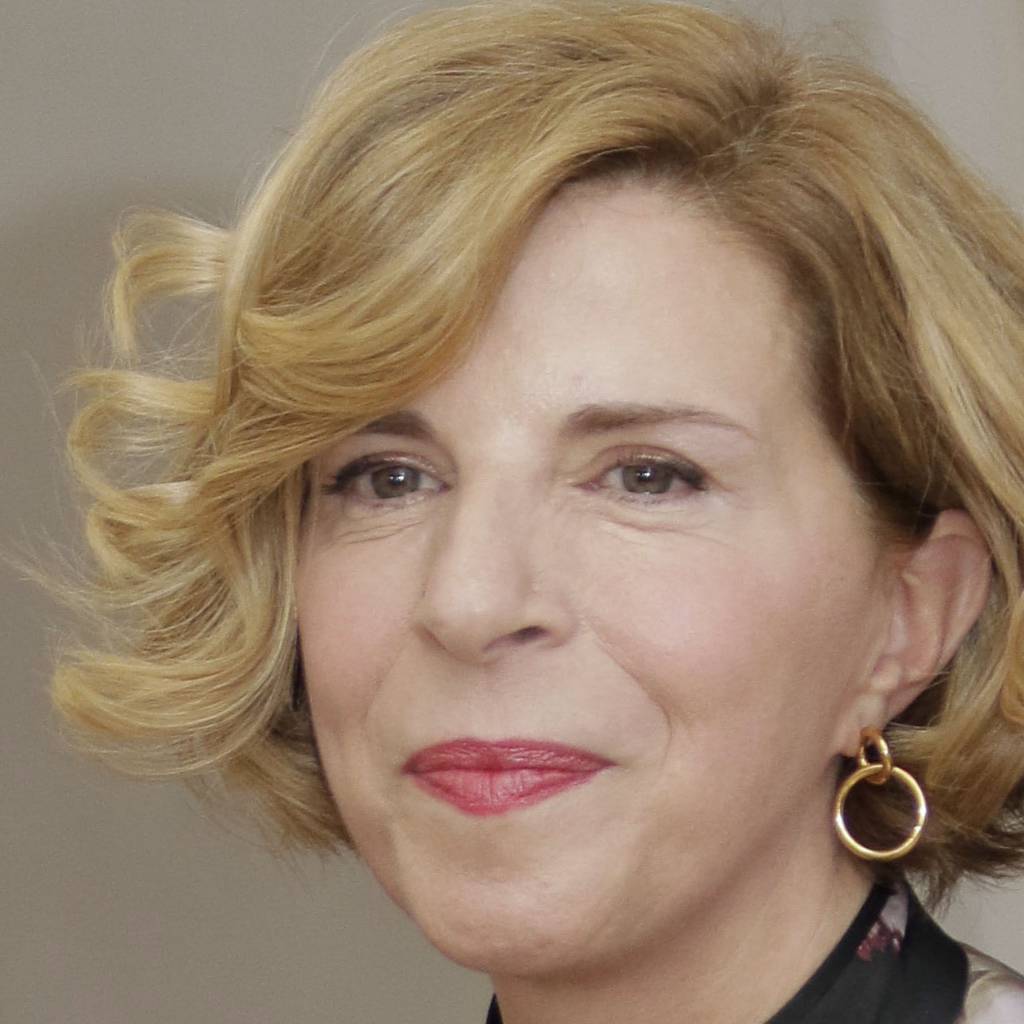}
    \end{subfigure}
    \begin{subfigure}[b]{0.24\linewidth}
        \includegraphics[width=\linewidth]{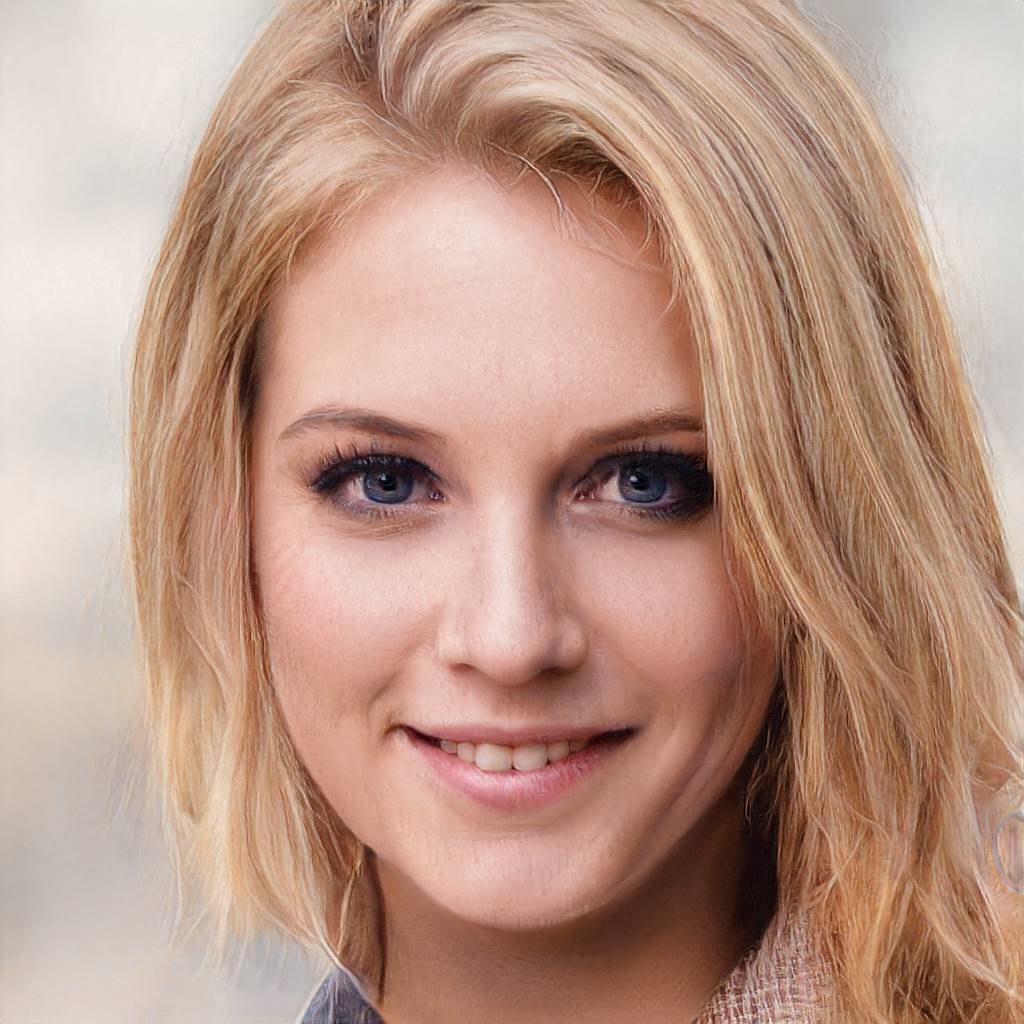}
    \end{subfigure}
    \begin{subfigure}[b]{0.24\linewidth}
        \includegraphics[width=\linewidth]{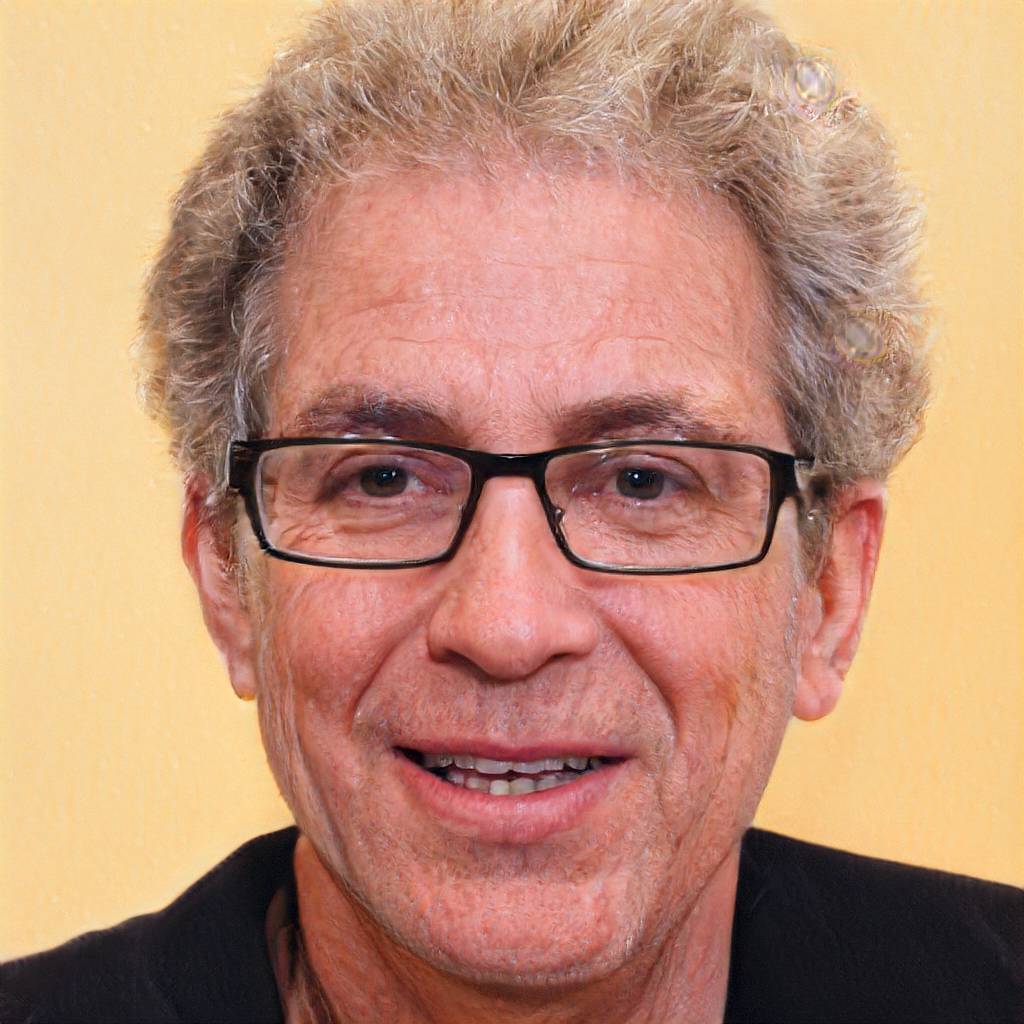}
    \end{subfigure}

    \vspace{0.2em}
    
    \begin{subfigure}[b]{0.24\linewidth}
        \includegraphics[width=\linewidth]{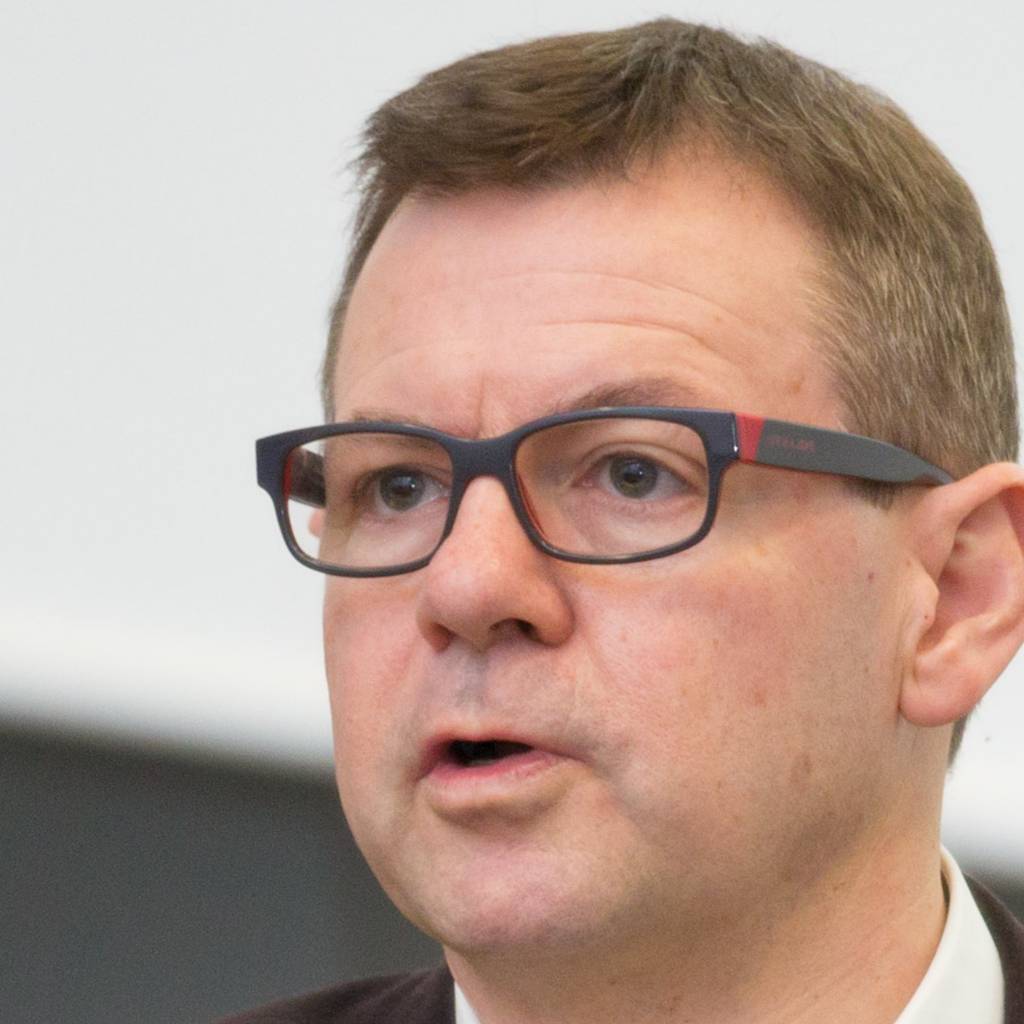}
    \end{subfigure}  
    \begin{subfigure}[b]{0.24\linewidth}
        \includegraphics[width=\linewidth]{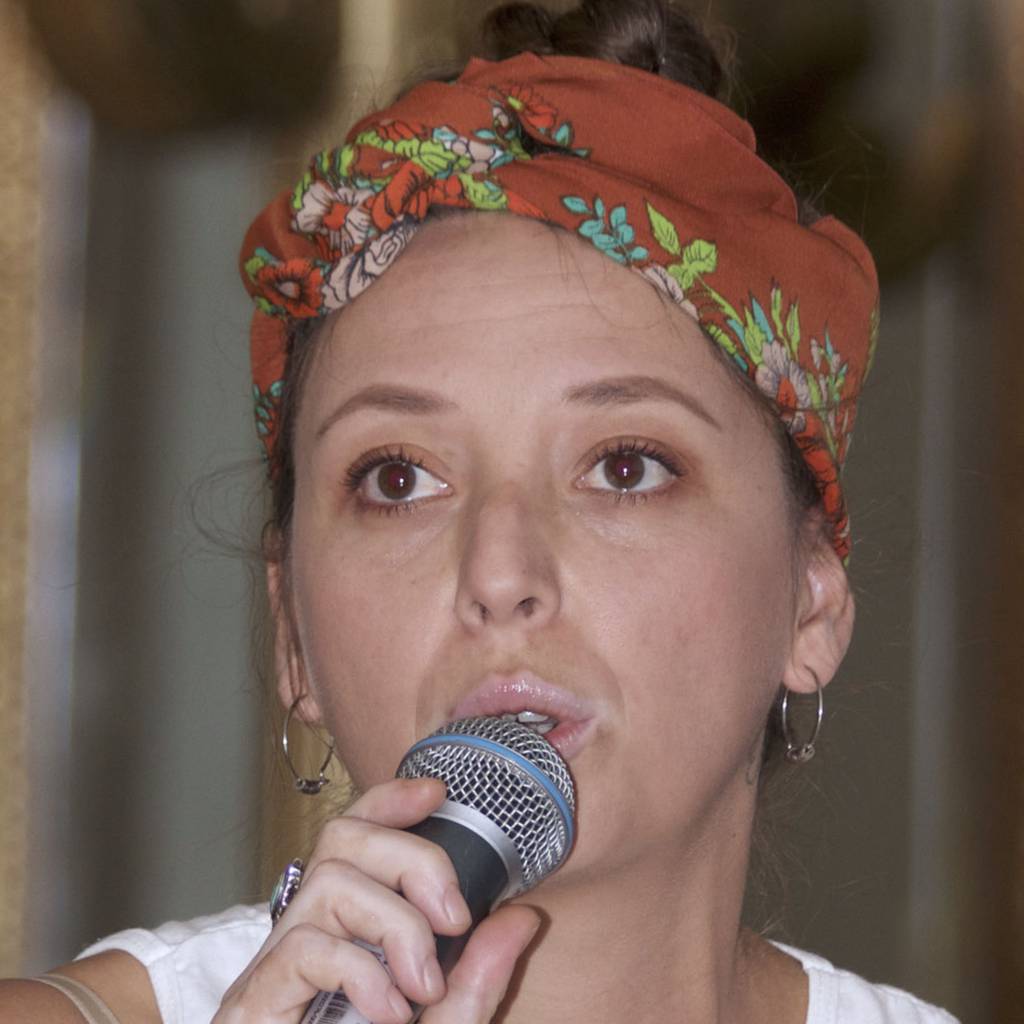}
    \end{subfigure}
    \begin{subfigure}[b]{0.24\linewidth}
        \includegraphics[width=\linewidth]{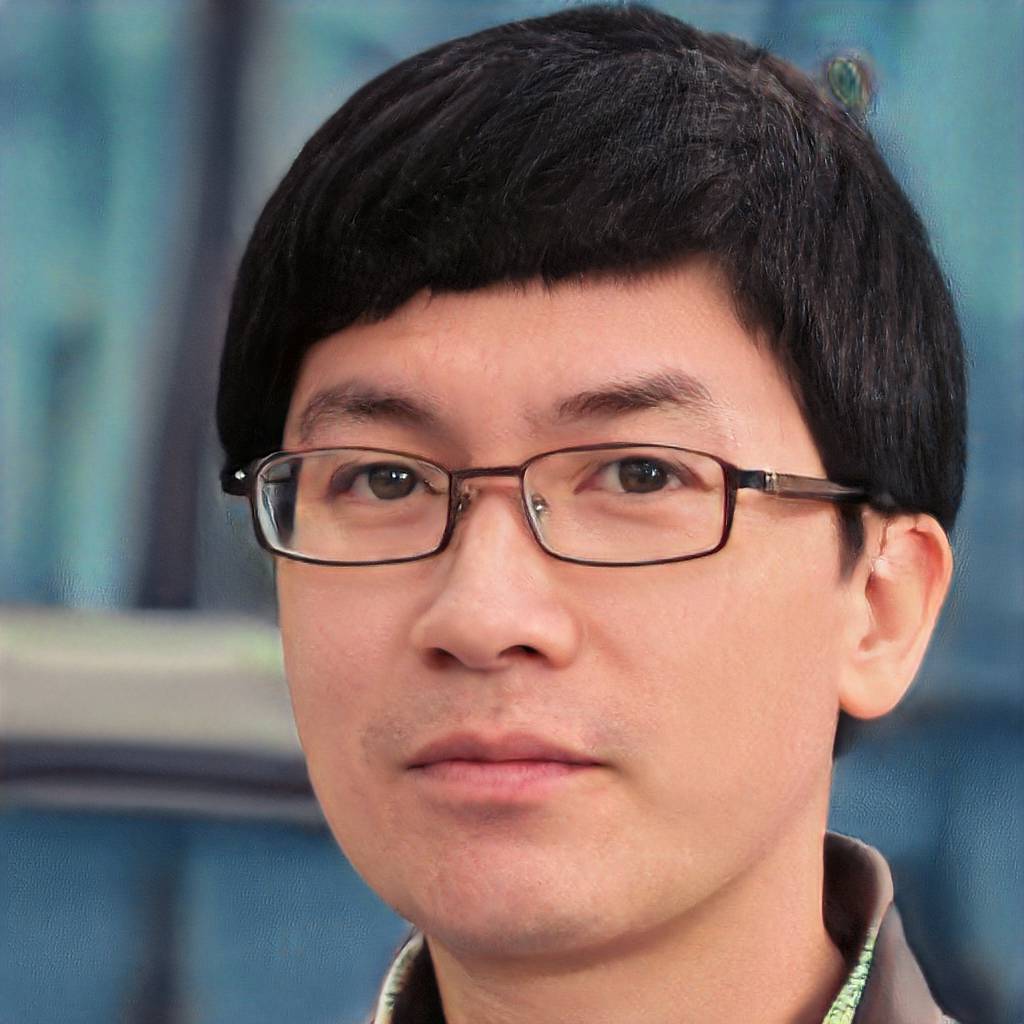}
    \end{subfigure}
    \begin{subfigure}[b]{0.24\linewidth}
        \includegraphics[width=\linewidth]{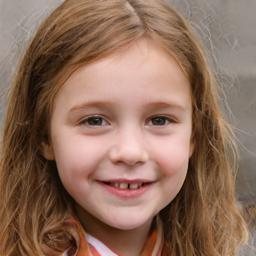}
    \end{subfigure}
    \caption{Can you distinguish fake from real images? The answers are shown below.\protect\footnotemark}
    \label{fig:example_ffhq_dataset}
\end{figure}
\footnotetext{Left four are \textit{real} (from the FFHQ dataset), and right four are \textit{generated} by StyleGAN (trained on the FFHQ dataset).}

Generative Adversarial Networks (GANs) \cite{goodfellow2014generative} could be regarded as the most promising and widely used type of generative models for image creation and manipulation. In only a few years of existence, many features such as: visual image quality; image resolution; range of control over the output; and ease of training these models have been improved. Recently, \cite{karras2019style} proposed StyleGAN, which is able to generate nearly photo-realistic facial images of 1024x1024 resolution, along with some stylistic control over the output, as presented in \Cref{fig:example_ffhq_dataset}. \cite{karras2019analyzing} has proposed an improved version with reduced visual artefacts.
To counteract the development of generative models, automatic fake imagery detection methods have gained increasing interest. Many works focus on learning-based detection, using Convolutional Neural Networks (CNNs). They work well on data similar to that seen during training, but often fail when images are generated by other GANs \cite{cozzolino2018forensictransfer} or when images are post-processed \cite{marra2018detection}. 

\begin{figure*}[!htbp]
\centering
\captionsetup{font=small}
\includegraphics[width=0.8\textwidth]{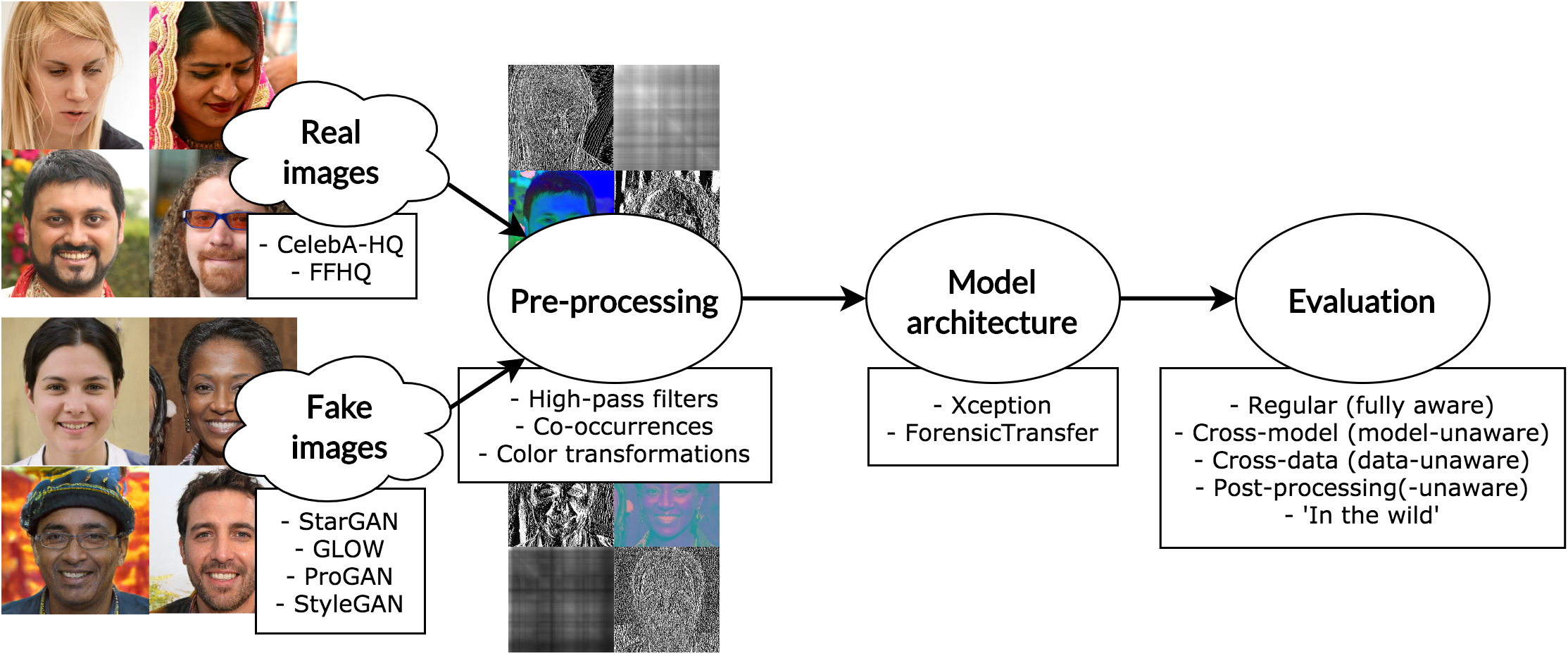}
\caption{Overview of our experimental pipeline. Two state-of-the-art detection models are evaluated under real-world scenarios with a focus on cross-model, cross-data and post-processing scenarios. Pre-processing techniques are examined for generalizability.}
\label{fig:pipeline}
\end{figure*}

Deviations in data sources and post-processing techniques are inconvenient in real-world scenarios. In this work, we refer to real-world scenarios as scenarios where an image encountered has an unknown source and possibly underwent unknown forms of post-processing after its creation. Furthermore, an image should be of reasonable size and should have no clearly visible alterations that lowers its credibility of being authentic. An example of such a real-world scenario is a forensic setting where the authenticity of an image must be determined. It is desirable that a detection method works well, independently of the type of model that generated or manipulated the encountered image. Another example is images encountered on social media pages, which may be unintentionally or deliberately altered. Examples of unintentional alterations are compression and resampling (or resizing), which often happen when uploading images onto social media or viewing images in a web browser. Blurring, adding noise, and adjusting colors are examples of deliberate alterations. We also assume that in real-world scenarios, the majority of users that encounter these images are not trained to detect fake images. Based on the trend of applications using DeepFakes and the advanced techniques to create realistic fully-generated images, we expect that the use of fully-generated images will be the next trend for new applications. For this reason, it is important to take both real-world conditions and fully-generated images into consideration when evaluating detection methods.

In this work, we aim to evaluate state-of-the-art image generation models under an approximation of real-world conditions, using the following three categories: 1) a \textit{cross-model scenario}, where the type of model used to generate an image is unknown, 2) a \textit{cross-data scenario}, where the data used to train a generative model is unknown, and 3) a \textit{post-processing scenario}, where an image is modified with an unknown type of post-processing. For each category we examine whether the generalizability of learning-based methods could be improved using commonly used pre-processing methods. Our work focuses on facial images, since most applications are targeted on facial generation or manipulation.

Our main contributions are the following: \\
1) We propose a framework, presented in \Cref{fig:pipeline}, consisting of three types of evaluation required for robust evaluation under real-world conditions: cross-model, cross-data, and post-processing evaluation;\\ 
2) We evaluate the most promising state-of-the-art model architectures and pre-processing methods;\\
3) We perform a user study with 496 participants and measure human performance of detecting state-of-the-art generated images and factors that influence this performance.

%% file: sections/related_work.tex
\section{Related work}
In this section, we review methods for CNN-generated image detection, image pre-processing and human detection of image forgery. GANs \cite{goodfellow2014generative} have recently emerged as the state-of-the-art in generating realistic imagery, in terms of image resolution and visual quality. Recent works have been able to generate nearly photo-realistic facial images \cite{karras2019style,karras2019analyzing}. Other works focus on more control over the output images, mainly in the fields of stylistic manipulation \cite{choi2018stargan,isola2017image,zhu2017unpaired} or semantic manipulation \cite{iizuka2017globally,park2019semantic,pathak2016context,yeh2017semantic}. Our work includes models capable of unconditional generation \cite{karras2017progressive,karras2019style} and conditional stylistic manipulation \cite{choi2018stargan,karras2019style,kingma2018glow} of human faces.

\noindent\textbf{CNN-generated image detection}
Early work of CNN-generated image detection uses handcrafted features based on domain knowledge. Two examples of domain knowledge that could be exploited are image color information and human facial appearances \cite{li2018detection,matern2019exploiting,mccloskey2018detecting,yang2019exposing}.
While these methods have reasonable performance, such handcrafted features are less applicable in real-world scenarios, where images often do not adhere to some of the assumptions made for these methods \eg when faces are partially covered. In this case, the methods of \cite{matern2019exploiting,yang2019exposing} might not work. 

Following these works, learning-based methods have been proposed, using CNNs to automatically learn features of real and generated images \cite{afchar2018mesonet,bayar2016deep,chollet2017xception,cozzolino2017recasting,cozzolino2018forensictransfer, rahmouni2017distinguishing}. \cite{cozzolino2018forensictransfer} presents ForensicTransfer and achieves state-of-the-art results on detecting CNN-inpainted images \cite{iizuka2017globally,yu2018generative} and fully CNN-generated images \cite{choi2018stargan,jin2017cyclegan,karras2017progressive,kingma2018glow}.
Another commonly used architecture for CNN-generated image detection is the Xception model \cite{chollet2017xception}, originally proposed as image classification model trained on ImageNet \cite{russakovsky2015imagenet}. \cite{marra2018detection,rossler2019faceforensics++} both evaluate several models and show that Xception yields the best overall performance across regular and compressed images in detecting fully-generated images \cite{marra2018detection} and CNN-manipulated images \cite{rossler2019faceforensics++}. Futhermore, the evaluation of  \cite{cozzolino2018forensictransfer} shows good performance of Xception, in some evaluation setups outperforming the ForensicTransfer model.
\cite{wang2019cnn} proposes a model, along with several data augmentation procedures, to detect fully-generated images of unknown sources. The results suggest that increasing the number of image classes, as well as randomly blurring and compressing images during training, increases the robustness of CNN-based detectors, yielding good results in cross-model and post-processing scenarios.
\cite{liu2020global} finds that real and fake images have textural differences and exploit this by proposing a \textit{Gram-Net} model architecture to focus on global image textures, yielding good results in cross-model, cross-data, and post-processing scenarios. 

We select ForensicTransfer \cite{cozzolino2018forensictransfer} and Xception \cite{chollet2017xception} for our evaluation. We did not take the architectures of \cite{liu2020global} and \cite{wang2019cnn} into account, since they were not published yet at the time this research was conducted. Given that these works are extensions of the state-of-the-art models we expect our results are still valid since our work focuses on different types of pre-processing techniques and datasets. We also evaluate an \textit{in the wild} scenario exclusively for facial images. Additionally, we are the first to perform a large scale user study that compare human performance under realistic conditions to model performance. 

\noindent\textbf{Image pre-processing}
\label{relatedwork_pp}
Pre-processing an image before passing it to a CNN-based model is not uncommon in the field of image forgery detection and has been studied by several works \cite{chen2015median,cozzolino2014image,cozzolino2017recasting,cozzolino2018forensictransfer,li2018detection, li2018identification,nataraj2019detecting,rao2016deep}. The motivation is to enrich or focus on specific information in the image, such that learning the difference between real and generated (or manipulated) might be more fruitful. As shown by \cite{albright2019source,marra2019gans,yu2019attributing}, CNN-generated images have pixel patterns dissimilar to real images, which might become more distinctive by learning more intrinsic (pixel-level) image features, such that detection models might generalize better to unseen (\eg model-unaware) fake images.

Several works on image forgery detection \cite{cozzolino2017recasting,cozzolino2018forensictransfer,li2018identification,mo2018fake,rao2016deep} include high-pass filters as a way to accentuate the high-frequency structure of an image.
Another type of pre-processing is color transformation, where non-RGB color information is used to detect forgeries. \cite{li2018detection} has shown the effectiveness of detecting generated images, using HSV (hue, saturation, value) and YCbCr (luma, red-chroma difference, and blue-chroma difference) color information along with a feature-based approach. Lastly, several works use co-occurrence matrices to focus on irregularities in pixel-patterns, for example in steganalysis \cite{cozzolino2017recasting,fridrich2012rich,pevny2010steganalysis,sullivan2006steganalysis,sullivan2005steganalysis} and detection of forged images \cite{cozzolino2014image,cozzolino2017recasting}. Recently, \cite{nataraj2019detecting} has used this approach for detecting CNN-generated images, suggesting good performance in several evaluation scenarios. 
Most works seem to evaluate one type or class of pre-processing method(s) with one model architecture \cite{cozzolino2018forensictransfer,mo2018fake,nataraj2019detecting}. The interaction between pre-processing methods and model architectures remains unclear as well as the benefits of pre-processing methods. In our work, we focus on these interactions by examining three common types of pre-processing: 1) high-pass filters, 2) co-occurrence matrices, and 3) color transformations.\\
\noindent\textbf{Human detection of image forgery}
Humans have trouble distinguishing forged images from authentic images, especially when no comparison material is provided to them \cite{nightingale2017can,schetinger2017humans,zheng2019survey}. Examples include detection of erase-fill, copy-move, cut-paste, and changes in reflections. \cite{rossler2019faceforensics++} shows that humans have trouble detecting CNN-modified images.

Recent work by \cite{zhou2019hype} addresses human performance on fully-generated GAN-images specifically. However, their aim is to evaluate the quality of GAN-images, not the human detection capabilities. Their results show that StyleGAN images generated using the \textit{truncation trick} are perceived as more realistic \cite{zhou2019hype}. The truncation trick refers to how far away a latent style vector is sampled from the average latent style vector, which determines the amount of variety in the generated image. Furthermore, images of 64x64 resolution are harder to distinguish from real than 1024x1024 images. However, images of this small size do not occur often in real-world scenarios.
Lastly, \cite{liu2020global} examines human performance of detecting GAN-generated images as a direct comparison with algorithmic detection. Therefore, they train humans by showing many examples, and then test them with novel examples, resulting in an average classification score of 63.9\% for the FFHQ vs StyleGAN\textsubscript{{\scriptsize \textit{FFHQ}}} scenario. While this yields an indication for upper bound performance of humans, it does not examine performance of untrained humans, and factors that influence performance, making it difficult to project the results to real-world scenarios.

This work attempts to determine human performance under an approximation of real-world conditions. It differs from \cite{liu2020global} since we do not pre-train participants, and measure the performance related to intermediate feedback. Moreover, it differs from \cite{zhou2019hype} since we do not include any time constraints or training phase and evaluate more logical image resolutions. Lastly, we examine the influence of AI-experience on human performance, and image cues humans use to recognise generated images.

%% file: sections/methods.tex
\section{Methods \& Experimental Setup}

\Cref{fig:pipeline} gives an overview of our method. Each component will be discussed next.

\subsection{Datasets}

\noindent\textbf{Real images} CelebA-HQ (CAHQ) \cite{karras2017progressive} and Flickr-Faces-HQ (FFHQ) \cite{karras2019style} are selected as datasets for real images. The first is a high-quality version of the original CelebA dataset \cite{liu2018large}, consisting of 30K front view facial pictures of celebrities. Note that high-quality refers to several processing steps as discussed by \cite{karras2017progressive}, yielding high-resolution and visually appealing images. The second is a dataset with 70K high-quality front view pictures of ordinary people, of which the first 30K are selected.

\noindent\textbf{Generated (fake) images} We use five datasets of generated images for evaluation under real-world conditions: 1) StarGAN\textsubscript{{\scriptsize \textit{CAHQ}}} \cite{choi2018stargan}, 2) GLOW\textsubscript{{\scriptsize \textit{CAHQ}}} \cite{kingma2018glow}, 3) ProGAN\textsubscript{{\scriptsize \textit{CAHQ}}} \cite{karras2017progressive}, 4) StyleGAN\textsubscript{{\scriptsize \textit{CAHQ}}} \cite{karras2019style}, and 5) StyleGAN\textsubscript{{\scriptsize \textit{FFHQ}}} \cite{karras2019style}.

The first two datasets are provided by \cite{cozzolino2018forensictransfer}. StarGAN and GLOW are conditional generative models that transform the style of an input image to some desired style. The datasets are created by taking a CAHQ image as input, randomly selecting a facial attribute out of a small set of attributes (\eg hair color), and generating the corresponding image with either the StarGAN or GLOW model. GLOW is not a GAN but a flow-based deep generative model. The third dataset consists of images generated by ProGAN, an unconditional GAN that generates high-resolution facial images. We use the dataset provided by \cite{karras2017progressive}. 

For the last two datasets, we use images generated by StyleGAN. StyleGAN could be regarded as the state-of-the-art GAN in terms of visual quality \cite{zhou2019hype}, strengthened by high-resolution images and some stylistic control over the output. We use two variants of StyleGAN images to evaluate cross-data performance. For the first variant, we use the dataset by \cite{karras2019style}. From the available sets of images generated with different amounts of truncation, we select the set generated using $\psi=0.5$. Note that these images are generated by a model trained on FFHQ images. There is no public StyleGAN\textsubscript{{\scriptsize \textit{CAHQ}}} dataset, thus we generate images using a model pre-trained on CAHQ images (with $\psi=0.5$). The motivation for selecting $\psi=0.5$ and the creation of StyleGAN\textsubscript{{\scriptsize \textit{CAHQ}}} are discussed in further detail in Section A.1 of the supplementary material.

For each dataset, we use 30K images, split into training (70\%), validation (20\%), and test (10\%) sets. The amount of real and fake images seen during training and testing is equal. During training, images are rescaled to match the corresponding input layer size of both models.

\begin{figure}[!t]
    \centering
    \captionsetup{font=footnotesize}
    \captionsetup[subfigure]{labelformat=empty}
    \begin{subfigure}[b]{0.235\linewidth}
        \centering        
        \includegraphics[width=\linewidth]{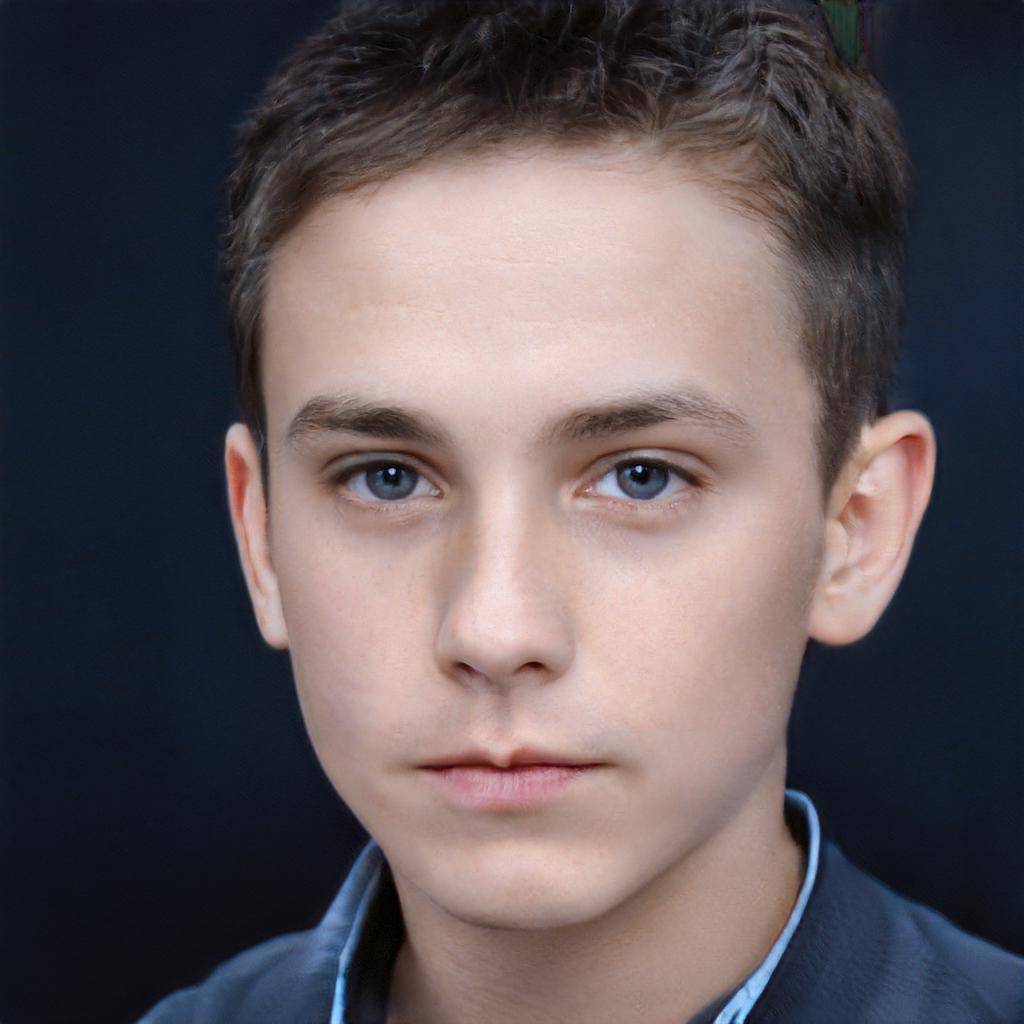}
        \caption{Regular image}
    \end{subfigure}
    \begin{subfigure}[b]{0.235\linewidth}
        \centering
        \includegraphics[width=\textwidth]{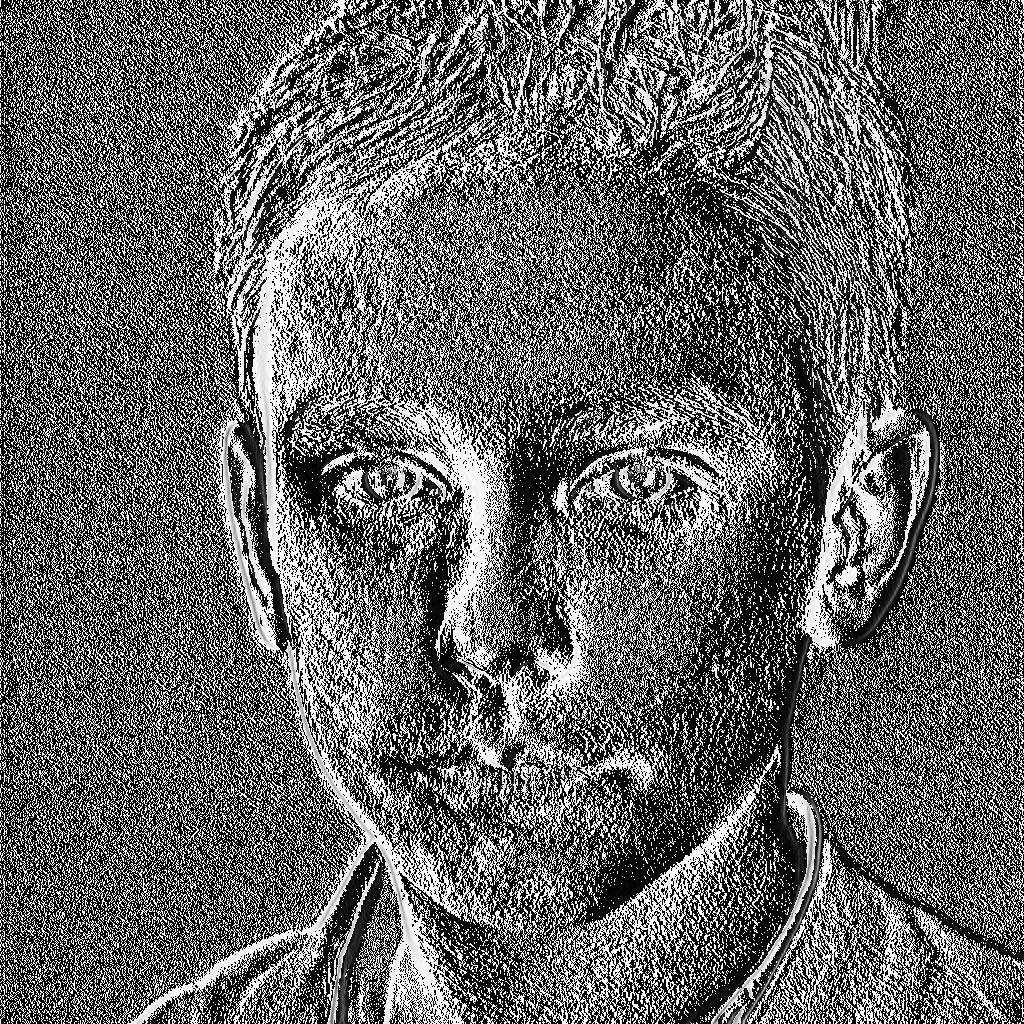}
        \caption{Res1 filtered}
    \end{subfigure}
    \begin{subfigure}[b]{0.235\linewidth}
        \centering
        \includegraphics[width=\textwidth]{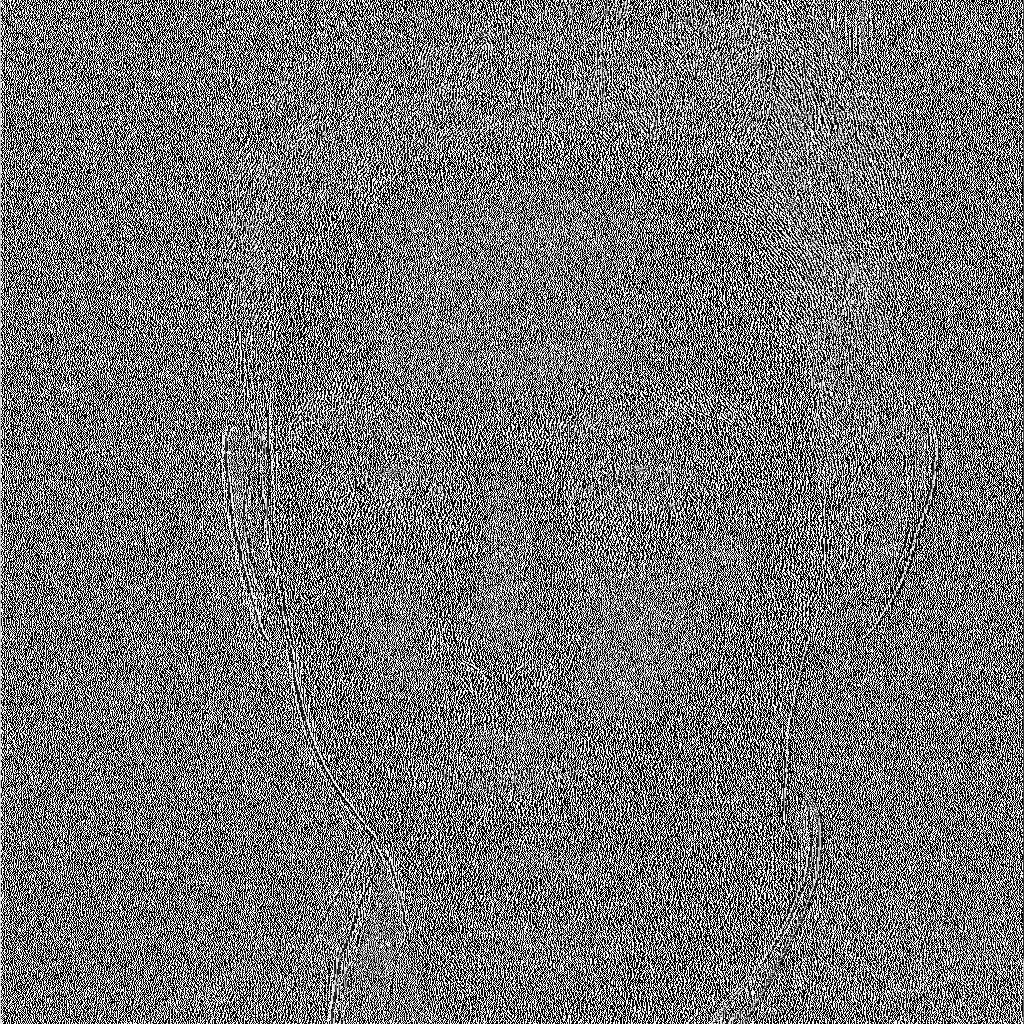}
        \caption{Res3 filtered}
    \end{subfigure}
    \begin{subfigure}[b]{0.235\linewidth}
        \centering
        \includegraphics[width=\textwidth]{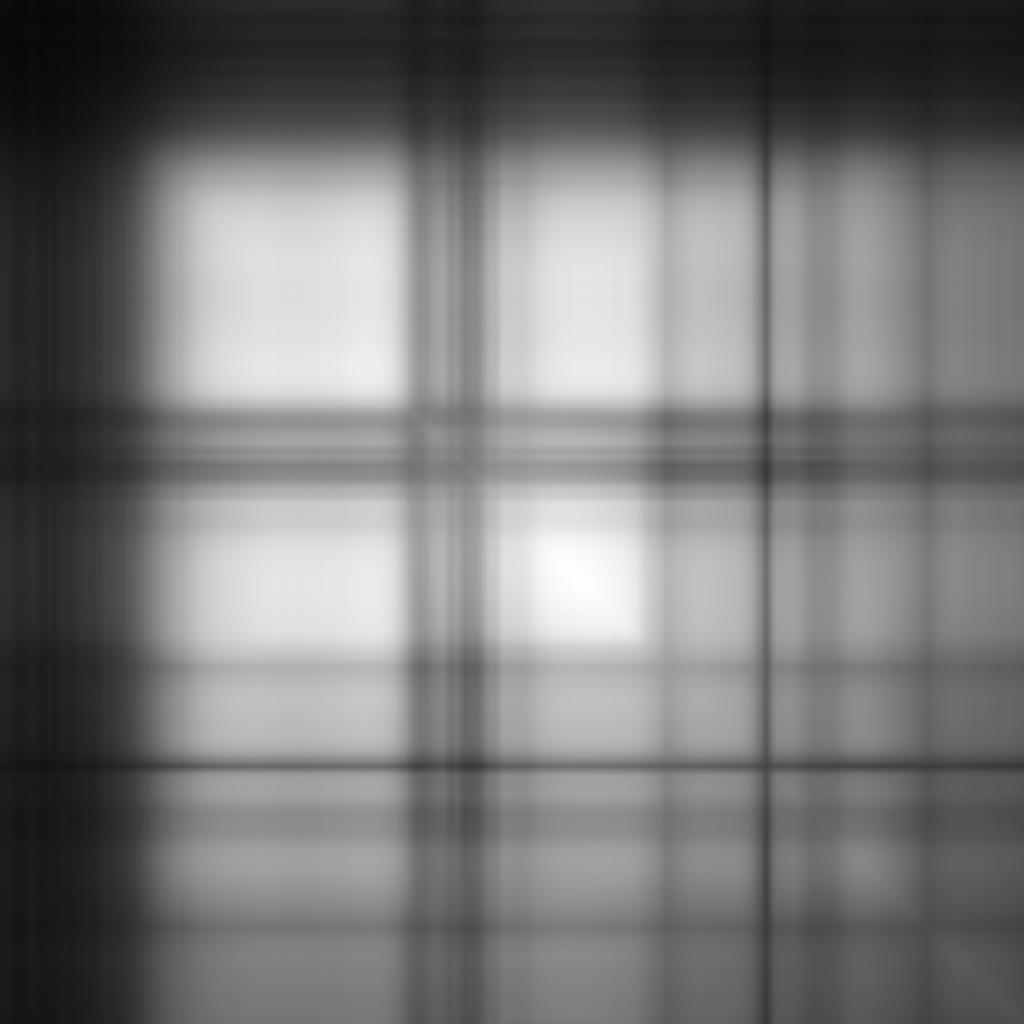}
        \caption{Cooc filtered}
    \end{subfigure}
    
    \caption{Visualization of several pre-processing methods using a StyleGAN\textsubscript{{\scriptsize \textit{FFHQ}}} image. Note that HSV is not visualized since it is not meaningful to display using RGB color conventions.}
    \label{fig:pp_visualization}
\end{figure}
%\hspace{1em}
\subsection{Pre-processing}
For pre-processing techniques we use high-pass filters, co-occurrence matrices, and color transformations, since these have recently been demonstrated to work well in CNN-generated image detection \cite{cozzolino2018forensictransfer,nataraj2019detecting,li2018detection}. For each of these three categories, one or multiple variants have been experimented with. We select the best performing methods to be included in the results. A visualization of these methods is shown in \Cref{fig:pp_visualization}. \textbf{Res1} is a first-order derivative filter: $[1 \hspace{0.90em} $\textminus$1]$ \cite{chen2017image,fridrich2012rich}. It is included as baseline high-pass filter. Similar to \cite{cozzolino2018forensictransfer}, this filter is applied in horizontal and vertical direction in parallel and the resulting channels are concatenated, yielding 6 image channels. \textbf{Res3} is a third-order derivative filter: $[1 \hspace{0.90em} $\textminus$3 \hspace{0.90em} 3 \hspace{0.90em} $\textminus$1]$ \cite{cozzolino2017recasting,cozzolino2018forensictransfer}. Again, it is applied similarly to Res1 and is equal to the \textit{RES} filter used by \cite{cozzolino2018forensictransfer}. Note that we have experimented with other implementations (\ie applying the filter horizontally and vertically in sequence, yielding three channels) but these performed worse, thus choosing the implementation by \cite{cozzolino2018forensictransfer}. \textbf{Cooc} calculates the co-occurrence matrix of an input image, similar to \cite{nataraj2019detecting}. This is done by a matrix multiplication of the original image with its transpose, resulting in three image channels. \textbf{HSV} converts to the hue, saturation, value (HSV) color space, resulting in three image channels. This is inspired by \cite{li2018detection}, who use HSV and YCbCr color spaces, as discussed in \Cref{relatedwork_pp}. Our initial experiments showed better performance of HSV, so YCbCr is not considered.

\subsection{Model architectures}
Based on the work of \cite{cozzolino2018forensictransfer,marra2018detection,rossler2019faceforensics++}, we select Xception \cite{chollet2017xception} and ForensicTransfer \cite{cozzolino2018forensictransfer} as state-of-the-art model architectures for CNN-generated image detection. \textbf{Xception (X)} \cite{chollet2017xception} is a deep CNN with depth-wise separable convolutions \cite{chollet2017xception}, inspired by Inception modules \cite{szegedy2015going}, and has shown good performance in multiple image forgery detection tasks \cite{cozzolino2018forensictransfer,marra2018detection,rossler2019faceforensics++}, both for regular and compressed images. \textbf{ForensicTransfer (FT)} is a CNN-based encoder-decoder architecture, which learns to encode the properties of fake and real images in latent space, outperforming several other methods when combined with high-pass filtering the images, or using transfer learning for few-shot adaptation to unknown classes \cite{cozzolino2018forensictransfer}. Images are classified as real if the \textit{real partition} in latent space is more active than the \textit{fake partition}, and vice versa. The training procedures for both models are described in Section A.2 of the supplementary material.

\subsection{Evaluation}
\label{subsec:evaluation}
To examine the performance of detection methods under real-world conditions, we include five types of evaluation.\\
\textbf{Default (fully aware)} In the easiest setup, test images are created by the same generative model as train images and are from the same data distribution. These test images are not further manipulated. This setup gives an upper bound on the performance of a detection method, but has no correspondence to a real-world scenario. We test this for StyleGAN\textsubscript{{\scriptsize \textit{CAHQ}}} and StyleGAN\textsubscript{{\scriptsize \textit{FFHQ}}}.\\
\textbf{Cross-model (model-unaware)} In a real-world scenario, many generative models exist and new models will be created in the future. In this setup, test images are generated by one or multiple different models than images in the training set. The detection model has no examples of similar test images. In our work, we evaluate the performance of 1) detecting StarGAN\textsubscript{{\scriptsize \textit{CAHQ}}}, GLOW\textsubscript{{\scriptsize \textit{CAHQ}}}, and ProGAN\textsubscript{{\scriptsize \textit{CAHQ}}} when trained on StyleGAN\textsubscript{{\scriptsize \textit{CAHQ}}}, and 2) detecting StyleGAN\textsubscript{{\scriptsize \textit{FFHQ}}} with $\psi \in [0.7, 1.0]$ when trained on $\psi=0.5$. \\
\textbf{Cross-data (data-unaware)} In a real-world setting, numerous different datasets could be used to train a generative model, each with their own biases and pre-processing methods, which have a large impact on the generated images. Thus, it is needed to evaluate how detection models can generalize to unknown images used for training a generative model. In this setup, the data used for generating training images differs from the data used for generating test images. The model may be equal or different.  In our work, we evaluate the performance of detecting StyleGAN\textsubscript{{\scriptsize \textit{FFHQ}}} test images when trained on StyleGAN\textsubscript{{\scriptsize \textit{CAHQ}}} images and vice versa.\\
\textbf{Post-processing(-unaware)} When images are uploaded to and downloaded from the internet, they are likely to undergo several types of post-processing, such as compression and resampling. On the other hand, images could be manipulated to make them less detectable, for example with blur and noise addition. In our work, we select two types of techniques, JPEG compression and Gaussian blurring, and evaluate how different amounts of post-processing influence the detection of StyleGAN\textsubscript{{\scriptsize \textit{FFHQ}}} images. We evaluate several degrees ranging from hardly visible to clearly visible to the human observer. \\
\textbf{In the wild} This mimics a real-world scenario where a detection model has access to all currently known state-of-the-art models and encounters images generated by a newer model. In our case, one detection model is trained on multiple known sources (StarGAN\textsubscript{{\scriptsize \textit{CAHQ}}}, GLOW\textsubscript{{\scriptsize \textit{CAHQ}}} and ProGAN\textsubscript{{\scriptsize \textit{CAHQ}}}), and evaluated on unknown sources of higher visual quality (StyleGAN\textsubscript{{\scriptsize \textit{CAHQ}}} and StyleGAN\textsubscript{{\scriptsize \textit{FFHQ}}}). 

%% file: sections/onlinesurvey.tex
\section{Online survey}
\begin{figure}[!t]
    \centering
    \captionsetup{font=footnotesize}
    \captionsetup{width=1.0\linewidth}
    \includegraphics[width=0.9\linewidth]{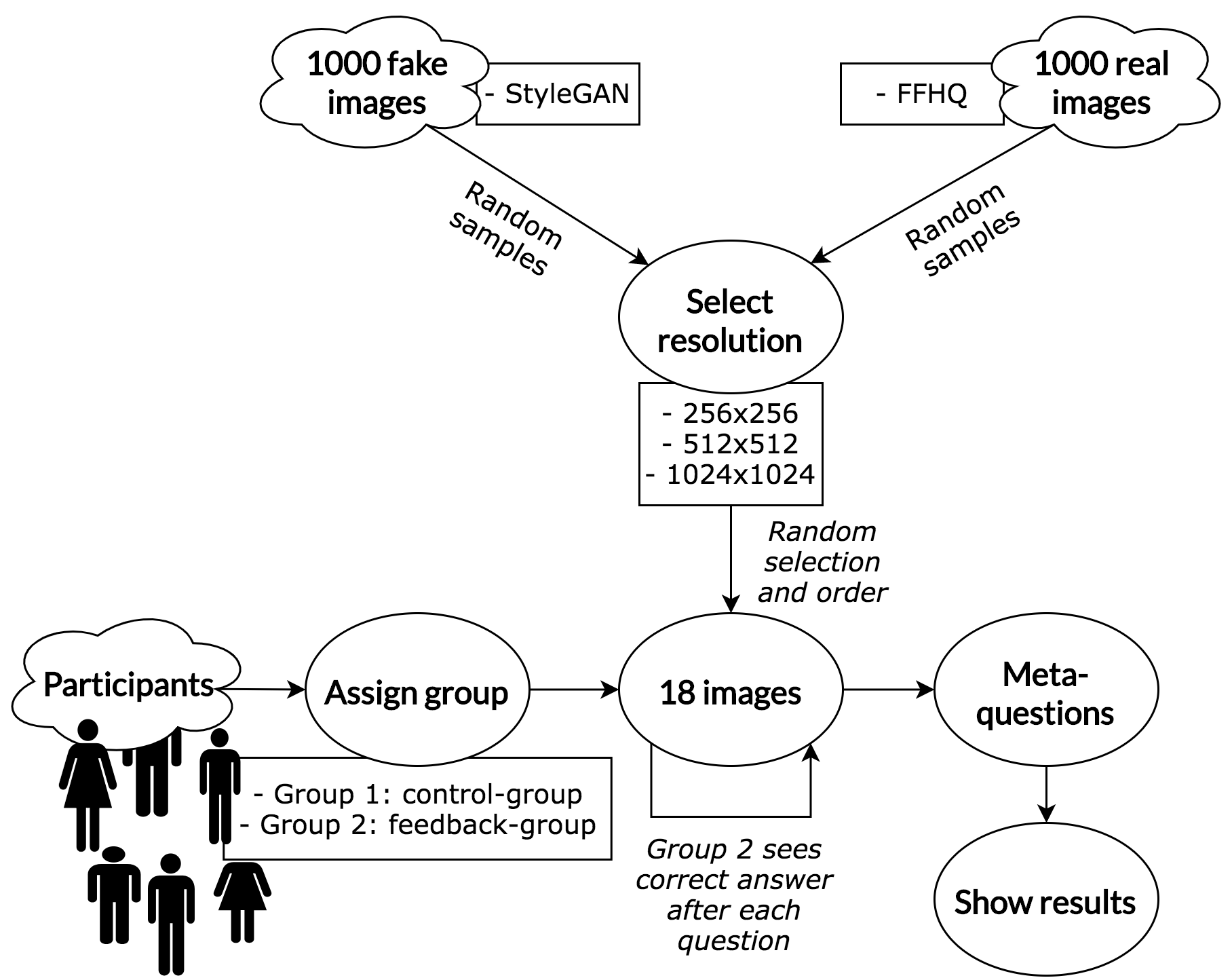}
    \caption[Schematic overview of online survey.]{Each participant is randomly assigned to the control-group or feedback group. Then he/she sees 18 images sequentially of varying resolutions and must decide for each image whether it is real or fake. }
    \label{fig:blokschema_survey}
\end{figure}

\begin{table*}[!htbp]
\scriptsize
\captionsetup{font=small}
\centering
\begin{threeparttable}
\tabcolsep=0.24cm
\begin{tabular}{l c c c c c c c c | c c c c c c}
\toprule
\multicolumn{2}{c}{Model}
        & \multicolumn{2}{c}{Default} 
                & \multicolumn{3}{c}{Cross-model} 
                            & \multicolumn{2}{c|}{Cross-data} 
                            & \multicolumn{2}{c}{Default}
                            & \multicolumn{2}{c}{Cross-model}
                            & \multicolumn{2}{c}{Cross-data}\\ % All \\ 
\cmidrule{1-15}

\multicolumn{1}{c}{\makecell{Pre-\\process}}
    & \multirow{2}{*}{Arch.}

        & \multicolumn{1}{c}{\makecell{StyleG\\{\tiny(CAHQ)}}}  
            & \textit{CAHQ} 
                & \multicolumn{1}{c}{\makecell{GLOW\\{\tiny(CAHQ)}}}
                    & \multicolumn{1}{c}{\makecell{ProG\\{\tiny(CAHQ)}}}
                    %& \multirowcell{2}{Pro\\GAN}

                        & \multicolumn{1}{c}{\makecell{StarG\\{\tiny(CAHQ)}}} 
                            & \multicolumn{1}{c}{\makecell{StyleG\\{\tiny(FFHQ)}}} 
                                & \textit{FFHQ} 
                                & \multicolumn{1}{c}{\makecell{StyleG\\{\tiny(FFHQ)}}}
                                & \textit{FFHQ} 
                                & 
                              \multicolumn{1}{c}{\makecell{StyleG\\$\psi=0.7$}} 
                    & \makecell{StyleG\\$\psi=1.0$} 
                        & \multicolumn{1}{c}{\makecell{StyleG\\{\tiny(CAHQ)}}} 
                                & \textit{CAHQ}
                                \\ %& \multicolumn{1}{|c}{Average}  \\
\cmidrule(r){1-2}
\cmidrule(lr){3-4}
\cmidrule(lr){5-7}
\cmidrule(lr){8-9}
\cmidrule(lr){10-11}
\cmidrule(lr){12-13}
\cmidrule(l){14-15}

% None
\multirow{2}{*}{\---}
    & X     & 99.6 & \textbf{99.8} & \phantom{0}0.3 & \phantom{0}1.0 & \phantom{0}0.2 & \phantom{0}5.9 & \textbf{99.8} & \textbf{99.9} & \textbf{100} & 97.3 & 84.1 & \phantom{0}0.2 & \textbf{100}\\ % 40.17 \\
    & FT    & 98.3 & 99.3 & 0.3 & \textbf{88.9} & 97.5 & 44.7 & 60.7 & 99.2 & \textbf{100} & 97.8 & \textbf{94.3} & \phantom{0}0.01 & \textbf{100}\\% 76.86 \\

% Res1
\multirow{2}{*}{Res1}
    & X     & 91.6 & 96.8 & \phantom{0}0.9 & 65.2 & 37.8 & 37.8 & 69.2 & 91.2 & \textbf{100} & 91.8 & 84.5 & \phantom{0}0.2 & 99.9\\ % 58.44 \\
    & FT    & 99.5 & 95.4 & 31.2 & \textbf{88.9} & \textbf{100} & \textbf{90.2} & 28.4 & 91.3 & 90.9 & 89.1 & 83.5 & \phantom{0}8.8 & 93.9\\% 82.99 \\

% Res3
\multirow{2}{*}{Res3}
    & X     & 72.2 & 45.6 & \textbf{49.9} & 65.7 & 57.8 & 62.1 & 39.8 & 54.5 & 98.6 & 54.4 & 51.2 & \phantom{0}2.5 & 98.7\\ %58.25 \\
    & FT    & 93.3 & 89.9 & 36.6 & 65.2 & 99.5 & 87.8 & 36.7 & 72.5 & 75.4 & 70.4 & 67.1 & \textbf{30.8} & 84.0\\% 76.91 \\

% Cooc
\multirow{2}{*}{Cooc}
    & X     & 95.0 & 96.4 & \phantom{0}2.1 & 12.9 & \phantom{0}2.8 & 31.2 & 95.3 & 93.6 & 97.4 & 63.5 & 18.6 & 13.7 & 98.7\\ % 41.84 \\
    & FT    & 79.6 & 77.8 & \phantom{0}2.3 & 37.3 & 26.8 & 32.4 & 83.8 & 80.4 & 91.5 & 52.9 & 23.7 & 20.6 & 91.8\\%  44.76 \\

% HSV
\multirow{2}{*}{HSV}
    & X     & \textbf{99.9} & \textbf{99.8} & \phantom{0}3.0 & 63.6 & 12.2 & 44.7 & 87.6 & \textbf{99.9} & 99.9 & \textbf{98.3} & 87.9 & \phantom{0}0.2 & 99.9\\ % 55.70 \\
    & FT    & 93.7 & 97.9 & 33.0 & 79.5 & 81.8 &  46.8 & 56.8 & 98.7 & 99.9 & 96.7 & 91.3 & \phantom{0}0.1 & \textbf{100}\\% 77.20 \\

\bottomrule
\end{tabular}
\caption{Evaluation of default, cross-model, and cross-data performance. The first setup uses StyleGAN\textsubscript{{\scriptsize \textit{CAHQ}}} ($\psi=0.5$) as a training dataset and tests on 1) StyleGAN\textsubscript{{\scriptsize \textit{CAHQ}}} images (default evaluation), 2) GLOW\textsubscript{{\scriptsize \textit{CAHQ}}}, ProGAN\textsubscript{{\scriptsize \textit{CAHQ}}} and StarGAN\textsubscript{{\scriptsize \textit{CAHQ}}} images (cross-model), and 3) StyleGAN\textsubscript{{\scriptsize \textit{FFHQ}}} images (cross-data). The second setup uses StyleGAN\textsubscript{{\scriptsize \textit{FFHQ}}} ($\psi=0.5$) as a training dataset and tests on 1) StyleGAN\textsubscript{{\scriptsize \textit{FFHQ}}} images (default evaluation), 2) StyleGAN\textsubscript{{\scriptsize \textit{FFHQ}}} ($\psi=0.7$ and $\psi=1.0$) images (cross-model), and 3) StyleGAN\textsubscript{{\scriptsize \textit{CAHQ}}} images (cross-data). On the left side, we denote the type of pre-processing and model architecture, where X denotes Xception and FT denotes ForensicTransfer. We have also abbreviated StyleG(AN)\textsubscript{{\scriptsize \textit{CAHQ}}}, ProG(AN)\textsubscript{{\scriptsize \textit{CAHQ}}}, and StarG(AN)\textsubscript{{\scriptsize \textit{CAHQ}}} for visualization purposes. Real image datasets are \textit{cursive} (CAHQ and FFHQ). Best accuracies per dataset (\ie column) are \textbf{bold}. Accuracies are averaged over 5 runs.}
\label{tab:cross_model_stylegan}
\end{threeparttable}
\end{table*}

To examine how well humans can identify state-of-the-art fake images, we conduct a user study with 496 participants. We also study what influences their performance. A schematic overview of the survey is given in \Cref{fig:blokschema_survey}. Each participant is randomly assigned to the control-group or feedback group. It then sees 18 images sequentially of varying resolutions and must decide for each image if it is real or fake. The whole process of survey design is described in Section B of the supplementary material.

Note that our experiments aim to estimate how well humans would perform in real-world scenarios, by examining several real-world factors that could influence this performance. These include 1) image resolution (measured with three resolutions), 2) how well people are trained (measured with a feedback and control group), and 3) AI-experience (measured with a question after completing the survey).

%% file: sections/results.tex
\section{Results}

\subsection{Algorithmic detection}

\Cref{tab:cross_model_stylegan} shows results of training on StyleGAN\textsubscript{{\scriptsize \textit{CAHQ}}} (left) and StyleGAN\textsubscript{{\scriptsize \textit{FFHQ}}} (right) images, along with cross-model and cross-data performance. It is important to note that we show the accuracy per dataset, and not the average accuracy of real and fake images combined. For this reason, the performance on the real and fake datasets do not add up to 100\%. For example, when a model is not able to detect fake images and classifies every image as real, the accuracy we report for the dataset with fake images will be 0\%. This is done to get a better understanding of how well each model can detect generated images, since real images are, with few exceptions, detected with high accuracies.\\
\textbf{Default (fully aware)} In the \textit{Default} columns of \Cref{tab:cross_model_stylegan} we see that both Xception and ForensicTransfer have a nearly perfect performance for both fake images (StyleGAN\textsubscript{{\scriptsize \textit{CAHQ}}}/StyleGAN\textsubscript{{\scriptsize \textit{FFHQ}}}) and real images (CAHQ/FFHQ). The third-order derivative filter seems to harm the performance the most for both model architectures and both datasets.\\ 
%This is the easiest setup, where test images come from the same generative model and dataset as training images. Moreover, test images are not further manipulated. This setup indicates the upper bound performance of a detection method, but has no correspondence to a real-world scenario where the data encountered is unknown.
\noindent\textbf{Cross-model (model-unaware)} 
For the cross-model setup, we see in \Cref{tab:cross_model_stylegan} that ForensicTransfer retrieves high performance for detecting ProGAN\textsubscript{{\scriptsize \textit{CAHQ}}} (88.9\%) and StarGAN\textsubscript{{\scriptsize \textit{CAHQ}}} (100\%) images. For ProGAN\textsubscript{{\scriptsize \textit{CAHQ}}} and StarGAN\textsubscript{{\scriptsize \textit{CAHQ}}}, the first order derivative filter yields slightly better results than using no filter. 
In our second setup, we train on StyleGAN\textsubscript{{\scriptsize \textit{FFHQ}}} and evaluate cross-model (parameter) evaluation. Note that this type of cross-model evaluation refers to the same model ($\psi=0.5$), but using another truncation ($\psi=0.7$ and $\psi=1.0$) for generating images, results in differences between training and testing images. The larger the difference between the $\psi$ values for training and testing, the lower the accuracy for detecting fake images. We also see this trend for different pre-processing techniques. For example, the performance for using the co-occurence matrix results in a drop of 45\% for Xception and 29\% for ForensicTransfer. \\ 
%In this setup, test images are generated by one or multiple different models than images in the training set. The detection model has no examples of similar test images. This is a logical setup in a real-world scenario, since many different generative models exist and will be created in the future. In our work, we evaluate the performance of detecting StarGAN, GLOW, and ProGAN when trained on StyleGAN\textsubscript{{\scriptsize \textit{CAHQ}}}.
\noindent\textbf{Cross-data (data-unaware)} 
In \Cref{tab:cross_model_stylegan} for both Xception as well as ForensicTransfer there is a trade-off between detecting fake and real images. Xception labels all images (99.8\%) as true when no pre-processing is used. In the same scenario, ForensicTransfer is able to detect fake StyleGAN\textsubscript{{\scriptsize \textit{FFHQ}}} images in 45\% of the cases, but as a consequence detecting FFHQ as real drops to 60\%. Using first or third order derivative filters for ForensicTransfer increases the performance for generated images, but decreases the performance for real images. %45\% to performances around 90\%.
For cross-data performance in our second setup, there is a an increase in performance of detecting StyleGAN\textsubscript{{\scriptsize \textit{CAHQ}}} images, when using ForensicTransfer together with third order derivative filters or the co-occurrence matrix. This increase from 0 to 20-30\% is still far from a good performance. 
%Here, the data used for generating training images differs from the data used for generating test images. The model may be equal or different. Similar to above, numerous different datasets could be used to train a generative model, each with their own biases and pre-processed methods, which have a large impact on the generated images. In our work, we evaluate the performance of detecting StyleGAN\textsubscript{{\scriptsize \textit{FFHQ}}} test images when trained on StyleGAN\textsubscript{{\scriptsize \textit{CAHQ}}} images.
Based on our results, there is no clear model or pre-processing method that stands out as best. ForensicTransfer has relatively high cross-model and cross-data performance, and seems to benefit slightly from high-pass filters, at the cost of a small drop in default performance. However, high-pass filters decrease performance for Xception, which seems to benefit slightly from HSV transformation. \\
\noindent\textbf{Post-processing(-unaware)} 
This evaluation consists of three levels of Gaussian blur, from a standard normal distribution with different kernel sizes, and three levels of JPEG compression using different quality factors. The results are presented in \Cref{tab:post_stylegan}. For each type of evaluation, the difference between training and testing images increases gradually to the right (\eg QF=90 is almost no compression, and QF=10 is severe compression). Without pre-processing techniques, Xception is much more robust to blur and compression, and shows nearly no drop in performance for the smallest amounts. For example, when using a 3x3 kernel for Gaussian blur Xception is able to detect StyleGAN\textsubscript{{\scriptsize \textit{FFHQ}}} images with 98.9\% accuracy, while ForensicTransfer only detects 18.1\% of the cases. Again, ForensicTransfer seems to benefit slightly from high-pass filters, while these deteriorate Xception performance. In this setup, HSV does not benefit Xception as much, making performance on cross-model and post-processing worse. Cooc shows no evident pattern in performance.
\begin{table}[!tp]
\scriptsize
\captionsetup{font=footnotesize}
\centering
\begin{threeparttable}
\tabcolsep=0.12cm
\begin{tabular}{l c c c c c c c c}
\toprule
\multicolumn{2}{c}{Model}
        & \multicolumn{1}{c}{\makecell{Default}}
                & \multicolumn{6}{c}{Post-processing} \\ 
\cmidrule{1-9}

\multicolumn{2}{c}{}
        & \multicolumn{1}{c}{} 
            & \multicolumn{3}{c}{Gaussian blur} 
                        & \multicolumn{3}{c}{JPEG compression} \\ 
\cmidrule(lr){4-6}
\cmidrule(l){7-9}

\makecell{Pre-\\proc.}
    & Arch.
        & \makecell{StyleG\\{\tiny (FFHQ)}}  
            & \makecell{3x3\\kernel} 
                & \makecell{9x9\\kernel} 
                    & \makecell{15x15\\kernel} 
                        & QF=90
                            & QF=50
                                & QF=10 \\
\cmidrule(r){1-2}
\cmidrule(lr){3-3}
\cmidrule(lr){4-6}
\cmidrule(l){7-9}

% None
\multirow{2}{*}{\---}
    & X         & \textbf{99.9} & \textbf{98.9} & \phantom{0}1.7 & \phantom{0}0.0 & \textbf{99.5} & \textbf{95.8} & 28.4 \\
    & FT        & 99.2 & 18.1 & \phantom{0}0.0 & \phantom{0}0.0 & \phantom{0}0.1 & \phantom{0}0.2 & \phantom{0}0.1 \\

% Res1
\multirow{2}{*}{Res1}
    & X         & 91.2 & 71.8 & \phantom{0}0.2 & \phantom{0}0.0 & 43.0 & \phantom{0}5.3 & \phantom{0}0.4 \\  
    & FT        & 91.3 & 79.5 & 21.9 & 12.4 & \phantom{0}8.0 & \phantom{0}1.4 & \phantom{0}0.9 \\

% Res3
\multirow{2}{*}{Res3}
    & X         & 54.5 & 11.0 & \phantom{0}0.8 & \phantom{0}1.3 & 16.9 & \phantom{0}6.3 & \phantom{0}5.7 \\
    & FT        & 72.5 & 66.4 & \textbf{65.7} & \textbf{64.1} & 17.7 & \phantom{0}5.9 & \phantom{0}3.8 \\
    
% Cooc
\multirow{2}{*}{Cooc}
    & X         & 93.6 & 92.3 & 39.6 & \phantom{0}0.8 & 93.0 & 91.9 & \textbf{75.5} \\  
    & FT        & 80.4 & 79.1 & 60.4 & 56.1 & 72.9 & 51.3 & \phantom{0}6.4 \\

% HSV
\multirow{2}{*}{HSV}
    & X         & \textbf{99.9} & 91.9 & \phantom{0}1.9 & \phantom{0}0.1 & 79.0 & 55.3 & 11.2 \\ 
    & FT        & 98.7 & 17.7 & \phantom{0}0.0 & \phantom{0}0.0 & \phantom{0}0.1 & \phantom{0}0.0 & \phantom{0}0.0 \\

\bottomrule
\end{tabular}
\caption{Evaluation of post-processing evaluation techniques using StyleGAN\textsubscript{{\scriptsize \textit{FFHQ}}} as a training dataset and testing on 1) StyleGAN\textsubscript{{\scriptsize \textit{FFHQ}}} images (default), 2) StyleGAN\textsubscript{{\scriptsize \textit{FFHQ}}} images with different amounts of Gaussian blur (three kernel sizes), and 3) StyleGAN\textsubscript{{\scriptsize \textit{FFHQ}}} images with different amounts of JPEG compression (three quality factors). The layout is similar to the previous table.}
\label{tab:post_stylegan}
\end{threeparttable}
\end{table}\\
\noindent\textbf{In the wild} As shown in \Cref{tab:in_the_wild}, cross-model detection is still low when training on images generated by different models. When examining the average accuracy of real images and unseen generated images (StyleGAN\textsubscript{{\scriptsize \textit{CAHQ}}}/StyleGAN\textsubscript{{\scriptsize \textit{FFHQ}}}), we observe that Xception without pre-processing performs best (62.4\%), followed by X-Cooc (61.2\%) and FT-Res1 (60.5\%). The other methods yield an average accuracy close to 50\% (with a balanced amount of real and generated images). Lastly, some pre-processing methods seem to decrease default performance.

\subsection{Human performance}
In \Cref{tab:feedback}, the results of our survey with 496 participants are presented. Note that in all tables, \textit{real} refers to an authentic image from the FFHQ dataset, while \textit{fake} refers to a generated image from the StyleGAN\textsubscript{{\scriptsize \textit{FFHQ}}} dataset. Out of all images, 70.1\% are labelled correctly. For real images, the average accuracy is 74.8\%, while for fake images it is 65.3\%. In the following, we examine the results of 1) intermediate feedback, 2) resolution, 3) AI-experience, and 4) upper and lower bound. The cues humans use to distinguish these images are analysed in Section C of the supplementary results.\\
% Influence of feedback
\noindent\textbf{Feedback} \Cref{tab:feedback} shows the average results of the group with intermediate feedback (N=233) and the group without (N=263). As shown, performance on real images is nearly identical, while performance on fake images is roughly 10\% higher, suggesting that participants can better learn to recognize fake images when receiving intermediate feedback. This is supported by an independent samples t-test, yielding a p-value of $< 0.005$ (with a t-statistic of 3.3). Note that only 18 images are evaluated in total, and this effect might be larger with more images. As a sanity check, the distribution of AI-experience among both groups is examined, which is nearly equal.
% 3.305

%Influence of resolution
\noindent\textbf{Image Resolution} When comparing performance with different image resolutions, \Cref{tab:feedback} shows that average detection accuracy of real and fake images decreases when images of lower resolution are presented. However, for real images this decrease is small, while for fake images the difference between highest and lowest resolution is 22.5\%. Note that each participant sees 3 real and 3 fake images of each resolution, but the selected images and order of presenting are completely random, excluding the possible influence of learning. The differences between resolution are tested with a one-way ANOVA test, yielding a p-value of $<< 0.001$ (with F-statistic 49.7). When performing post-hoc evaluation, we see that all group means differ much more than the standard error, suggesting that a lower image resolution makes an image significantly harder to classify, for the resolution tested in our survey. This is likely due to details and artefacts being less visible on smaller scales.

\begin{table}[!t]
\scriptsize
\captionsetup{font=footnotesize}
\centering
\begin{threeparttable}
\tabcolsep=0.10cm
\begin{tabular}{l c c c c c c c c c}
\toprule
\multicolumn{2}{c}{Model}
        & \multicolumn{5}{c}{Default} 
                & \multicolumn{3}{c}{\makecell{Cross-model}} \\
\cmidrule{1-10}

\makecell{Pre-\\proc.}
    & Arch.
        & \textit{FFHQ}
            & \textit{CAHQ}
                & \multicolumn{1}{c}{\makecell{StarG\\{\tiny(CAHQ)}}}
                    & \multicolumn{1}{c}{\makecell{GLOW\\{\tiny(CAHQ)}}}
                        & \multicolumn{1}{c}{\makecell{ProG\\{\tiny(CAHQ)}}}
                            & \makecell{StyleG\\{\tiny(CAHQ)}} 
                                & \makecell{StyleG\\{\tiny(FFHQ)}}
                                    & Avg\textbf{*} \\
\cmidrule(r){1-2}
\cmidrule(lr){3-7}
\cmidrule(l){8-10}
\multirow{2}{*}{\---}
    & X         & \textbf{99.9} & 99.9 & \textbf{100} & \textbf{100} & \textbf{99.7} & 49.5 & \phantom{0}0.1 & \textbf{62.4} \\
    & FT        & 89.3 & 99.1 & \textbf{100} & 99.9 & 78.8 & \phantom{0}7.9 & 10.4 & 51.7 \\

\multirow{2}{*}{Res1}
    & X         & 99.6 & 99.7 & 97.6 & 98.5 & 97.6 & \phantom{0}2.9 & \phantom{0}0.2 & 50.6 \\
    & FT        & 65.8 & 86.6 & \textbf{100} & \textbf{100} & 84.4 & 50.0 & 39.5 & 60.5 \\

\multirow{2}{*}{Res3}
    & X         & 43.0 & 41.0 & 83.5 & 81.6 & 78.8 & 52.8 & 58.0 & 48.7 \\
    & FT        & 49.7 & 43.6 & \textbf{100} & \textbf{100} & 76.3 & 76.3 & 45.6 & 53.8 \\

\multirow{2}{*}{Cooc}
    & X         & 97.3 & 96.4 & 99.2 & 99.6 & 94.3 & 50.1 & \phantom{0}0.8 & 61.2 \\
    & FT        & 27.6 & 15.7 & 88.4 & 85.7 & 86.8 & \textbf{86.3} & \textbf{69.3} & 49.7 \\

\multirow{2}{*}{HSV}
    & X         & \textbf{99.9} & \textbf{100} & \textbf{100} & \textbf{100} & \textbf{99.7} & 12.5 & 0.02 & 53.1 \\
    & FT        & 91.2 & 94.3 & \textbf{100} & \textbf{100} & 82.8 & 28.3 & \phantom{0}6.1 & 55.0 \\

\hline
\end{tabular}
\caption{Evaluation of 'in the wild' scenario. The models are trained on two datasets of real images (CAHQ and FFHQ) and three datasets of generated images (StarGAN\textsubscript{{\scriptsize \textit{CAHQ}}}, GLOW\textsubscript{{\scriptsize \textit{CAHQ}}}, and ProGAN\textsubscript{{\scriptsize \textit{CAHQ}}}). They are tested on two versions of StyleGAN images, that are not seen during training. The layout is similar to the previous tables. * Average of FFHQ, CAHQ, StyleGAN\textsubscript{{\scriptsize \textit{CAHQ}}} and StyleGAN\textsubscript{{\scriptsize \textit{FFHQ}}}, with an equal amount of real and generated images.}
\label{tab:in_the_wild}
\end{threeparttable}
\end{table}

% Overall results
\begin{table}[!t]
\scriptsize
\captionsetup{font=footnotesize}
\centering
\tabcolsep=0.15cm
\begin{tabular}{ccccccccc}
\toprule
    & \makecell{Total\\avg}
        & \multicolumn{2}{c}{\makecell{Intermediate\\Feedback}}
            & \multicolumn{3}{c}{\makecell{Image\\resolution}}
                & \multicolumn{2}{c}{AI-experience} \\
\cmidrule{1-9}
    &
        & No
            & Yes     
                & $1024^2$
                    & $512^2$
                        & $256^2$
                            & Little
                                & Much \\

\cmidrule(r){3-4}
\cmidrule(lr){5-7}
\cmidrule(l){8-9}

Real images     & 74.8 & 74.8 & 74.9 & 78.0 & 75.0 & 71.6 & 66.4 & 82.2 \\
Fake images     & 65.3 & 60.4 & 70.9 & 76.5 & 65.5 & 54.0 & 57.1 & 72.6 \\
All images      & 70.1 & 67.6 & 72.9 & 77.2 & 70.2 & 62.8 & 61.7 & 77.4 \\
\bottomrule
\end{tabular}
\caption{Average accuracies of labelling real and fake images among 1) all participants, 2) participants without/with intermediate feedback, 3) images of different resolution, and 4) participants with little/much AI-experience.}
\label{tab:feedback}
\end{table}

% Influence of AI\_experience
\noindent\textbf{AI-experience} \Cref{tab:feedback} shows the detection accuracies among two groups of participants with different levels of AI-experience. The first group (N=259) has \textit{much} AI-experience, and consists of AI-students, teachers, and professionals. The second group (N=218) has \textit{little} AI-experience and consists of all others. As shown, the average level of AI-experience within a group seems to have a large influence on detection performance. For real and fake images combined, the difference between little and much AI-experience is roughly 15\%. This difference is supported by an independent samples t-test, yielding a p-value of $<< 0.001$ (with a t-statistic of 10.7). Note that people with little AI-experience recognize fake images correctly in 57.1\% of the cases, which is slightly better than random.

\noindent\textbf{Upper and Lower Bound} The upper and lower bound of human performance is examined in \Cref{tab:upper_lower_bound}. This is done by evaluating the easiest scenario (\ie with feedback and 1024-res. images) and hardest scenario (\ie without feedback and 256-res. images). Within both scenarios, the difference between little and much AI-experience is examined. As becomes clear in \Cref{tab:upper_lower_bound}, the highest average detection accuracy for fake images is 86.7\% and the lowest is 37.0\%. 
\begin{table}[!t]
    \small
    \captionsetup{font=footnotesize}
    \centering
    \tabcolsep=0.11cm
    \begin{tabular}{ccccc}
    \toprule
        & \multicolumn{2}{c}{Upper bound}
            & \multicolumn{2}{c}{Lower bound} \\
    \cmidrule{1-5}
            & \makecell{Much AI\\experience}
                & \makecell{Little AI\\experience}
                    & \makecell{Much AI\\experience}
                        & \makecell{Little AI\\experience} \\
    \cmidrule(r){2-3}
    \cmidrule(l){4-5}
    Real images     & 85.4 & 69.6 & 78.9 & 61.0 \\
    Fake images     & 86.7 & 76.6 & 54.9 & 37.0 \\
    All images      & 86.0 & 73.1 & 66.9 & 49.0 \\
    \bottomrule
    \end{tabular}
    \caption{Average accuracies of labelling real and fake images in different setups, ranging from the most easy setup (left columns), which denote average performance of participants with feedback for 1024x1024 images, to the most difficult setup (right columns), which denote average performance of participants without feedback for 256x256 images. Within both groups, the performance of participants with little or much AI experience is shown.}
    \label{tab:upper_lower_bound}
\end{table}
\noindent\textbf{Comparison to algorithmic performance}
A comparison of algorithmic and human performance on StyleGAN\textsubscript{{\scriptsize \textit{FFHQ}}} data is presented in \Cref{fig:algorithmic_human_comparison}.
The \textit{upper bound} scenario approximates the most easy setup for both. For algorithmic detection this is the case when the model is trained and tested on the same dataset (StyleGAN\textsubscript{{\scriptsize \textit{FFHQ}}}). For humans this is the upper bound as shown in Table~\ref{tab:upper_lower_bound}. The \textit{realistic} scenario approximates real-world conditions. For algorithmic detection, we formulate this as the 'in the wild' scenario as shown in Table~\ref{tab:in_the_wild}, where only StyleGAN\textsubscript{{\scriptsize \textit{FFHQ}}} results are used. For humans it includes three variants (displayed from left to right in Figure~\ref{fig:algorithmic_human_comparison}):
1) an \textit{optimistic} realistic scenario, assuming humans have average AI-experience, learn to recognise fake images with feedback, and mainly see high-resolution images (512 and 1024),
2) an \textit{average} realistic scenario (estimated by the average of all survey results), and
3) a \textit{pessimistic} realistic scenario, assuming humans have low AI-experience, do not receive feedback, and see images of all resolutions.
Lastly, the \textit{lower bound} scenario presents the results of the most difficult setup. For humans this is the lower bound as shown in \ref{tab:upper_lower_bound}. For algorithmic detection, the lower bound is set at 50\%, which is a random guess in our two-class classification task with balanced class sizes. Note that its performance in the realistic scenario is already close to 50\%.

%% file: sections/conclusion.tex
\section{Conclusion \& Discussion}
Our work has evaluated two state-of-the-art models for detecting CNN-generated images, and has proposed three types of evaluation, along with an 'in the wild' setup, for mimicking real-world conditions in which such detection models will be used. Furthermore, we evaluated the benefits of several commonly used pre-processing methods. 

Based on our algorithmic experiments, we can conclude that performance in the easiest (default) scenario doesn't generalize well to other evaluation scenarios. ForensicTransfer seems more robust in cross-model performance, whereas Xception seems more robust in post-processing performance. Unfortunately, there is no single type of pre-processing that increases performance in multiple scenarios, and an increase in one evaluation setup is often paired with a decrease in other setups. Furthermore, the benefits of pre-processing methods are not guaranteed for both models; \ie high-pass filters work much better for ForensicTransfer than for Xception. Our results emphasize the importance of evaluating multiple scenarios. We emphasize the need for a benchmark dataset including images generated by multiple models, such that these types of evaluation can be performed and compared to related work.  

The results of the survey suggest that humans have trouble recognizing state-of-the-art fake images, which are correctly classified in roughly two-thirds of the cases. Our results suggest that the capability of detecting fake images could be influenced by several factors that may be of importance in real-world scenarios, such as AI-experience, image resolution, and feedback. When combining these factors, we see large differences between the best and the worst case (86.7\% as opposed to 37.0\% of fake images correctly recognized). These results emphasize the need for algorithmic detection methods to support humans in recognizing such images, as well as more research into the factors that influence human performance.
\begin{figure}[!t]
    \centering
    \captionsetup{font=footnotesize}
    \captionsetup{width=1.0\linewidth}
    \includegraphics[width=1.022\linewidth]{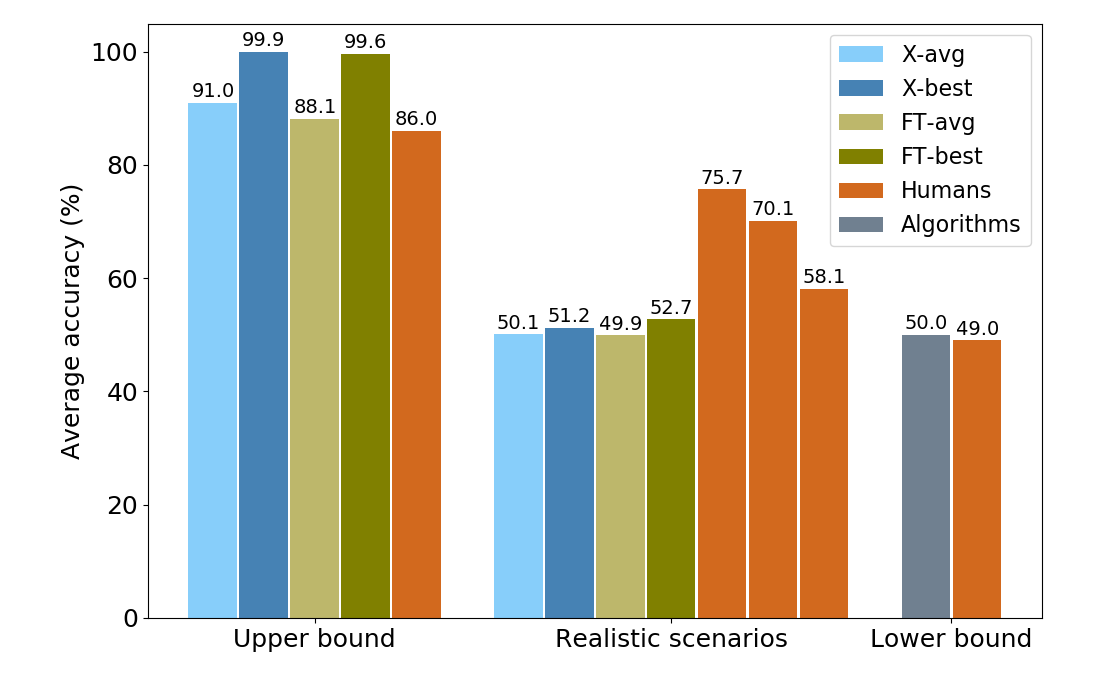}
    \caption{Comparison of algorithm and human performance in different scenarios.}
    \label{fig:algorithmic_human_comparison}
\end{figure}
Based on our comparison between algorithms and humans, we see that humans perform better than our models in the realistic scenario. However, from our upper bound performance we can conclude that models can outperform humans when trained and employed correctly. We encourage future work to pay more attention to extensiveness of evaluation which will result in more robust models for real-world scenarios.

%% file: sections/appendix.tex
\newpage

\def\thesection{\Alph{section}}
This supplementary material discusses more implementation details of our algorithmic pipeline (\Cref{sec:implementation_details}), additional information on how the survey is constructed (\Cref{sec:survey_design}), and insights about image cues that participants use to recognize fake images (\Cref{sec:image_cues}).

\section{Implementation Details}
\label{sec:implementation_details}

\subsection{Creation of StyleGAN\textsubscript{{\footnotesize \textit{\textbf{CAHQ}}}} dataset}
\label{subsec:stylegan_cahq}
\begin{figure}[!htbp]
    \centering
    \begin{subfigure}[b]{0.5\textwidth}
        \centering
        \includegraphics[width=0.22\linewidth]{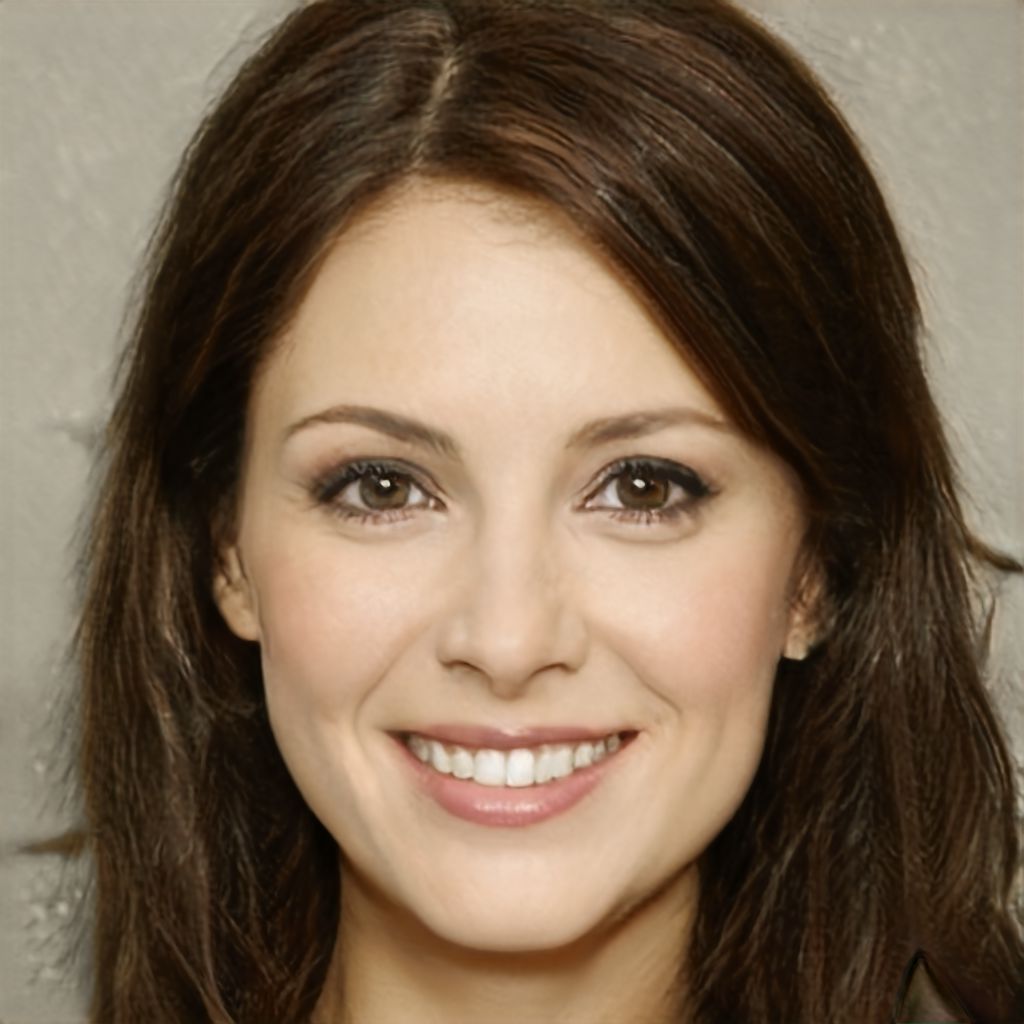}
        \includegraphics[width=0.22\linewidth]{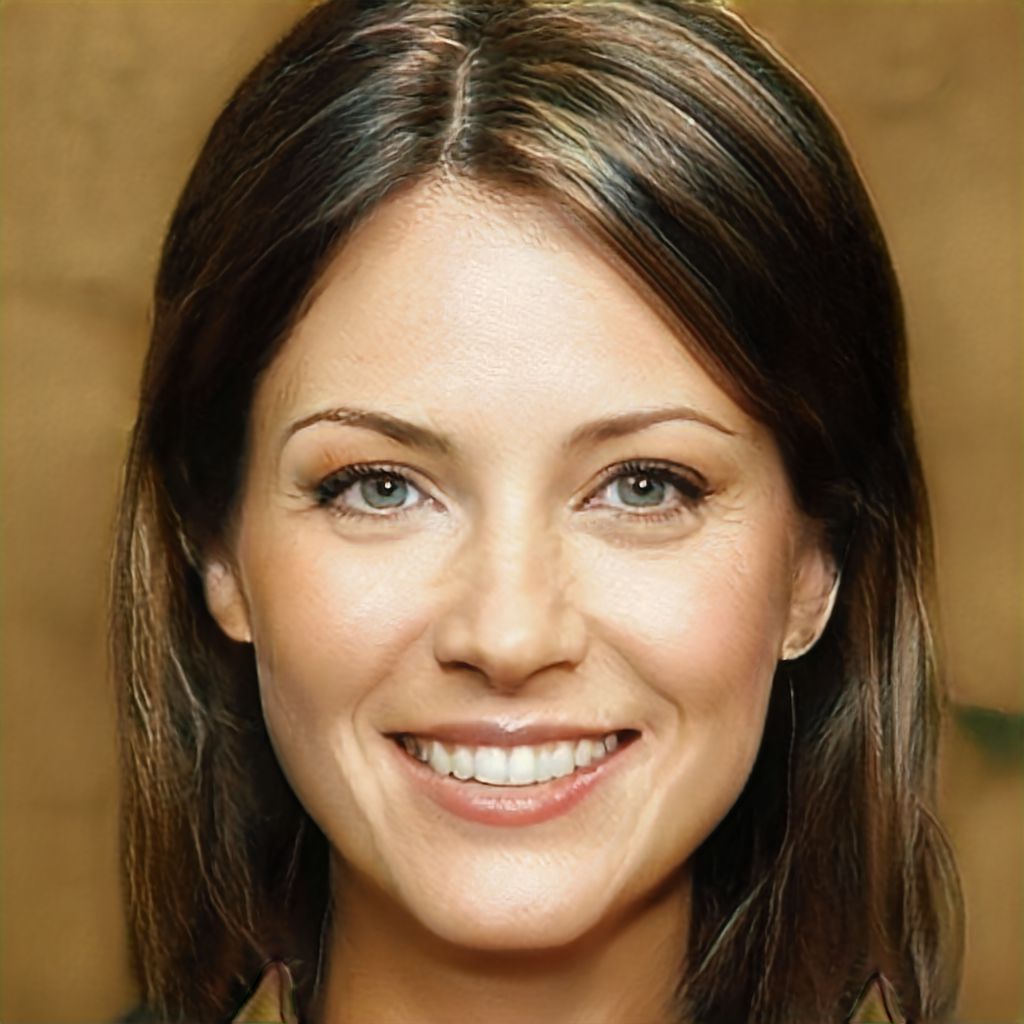}
        \includegraphics[width=0.22\linewidth]{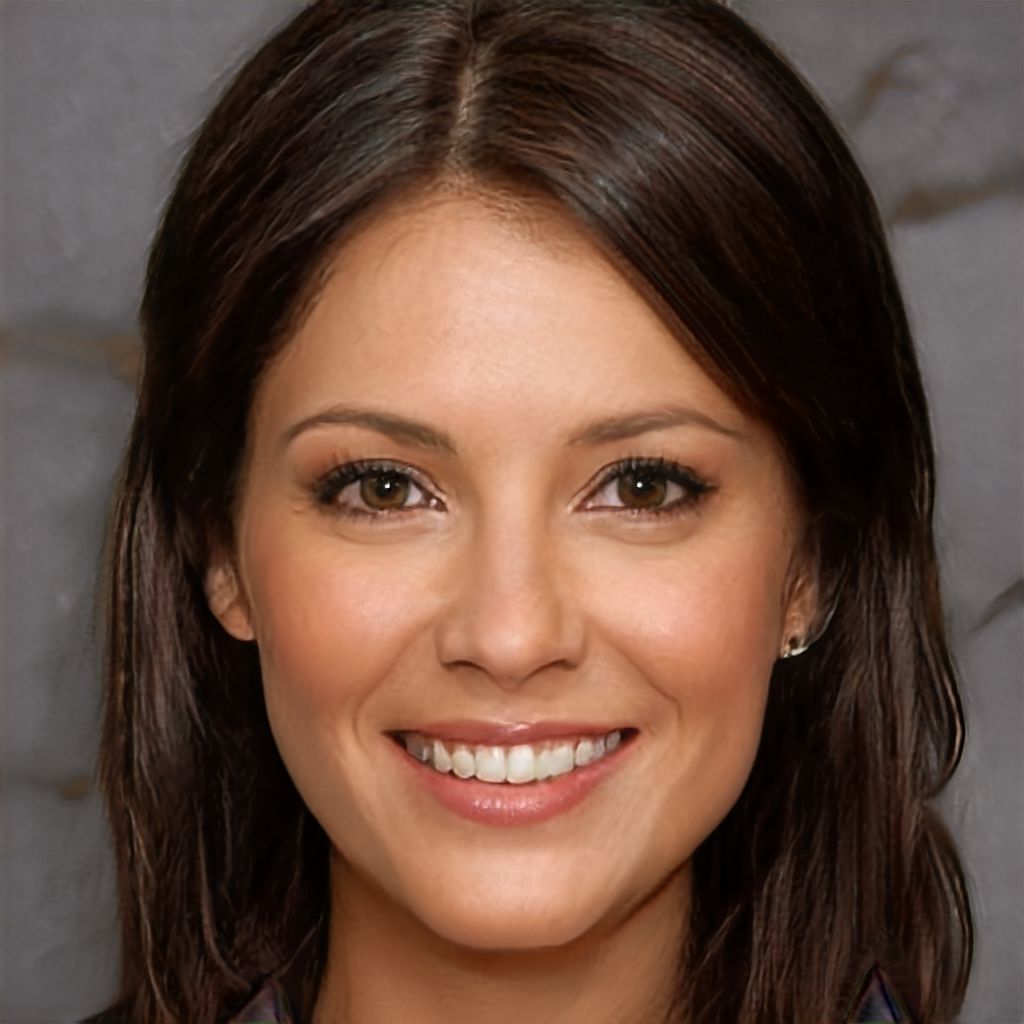}
        \includegraphics[width=0.22\linewidth]{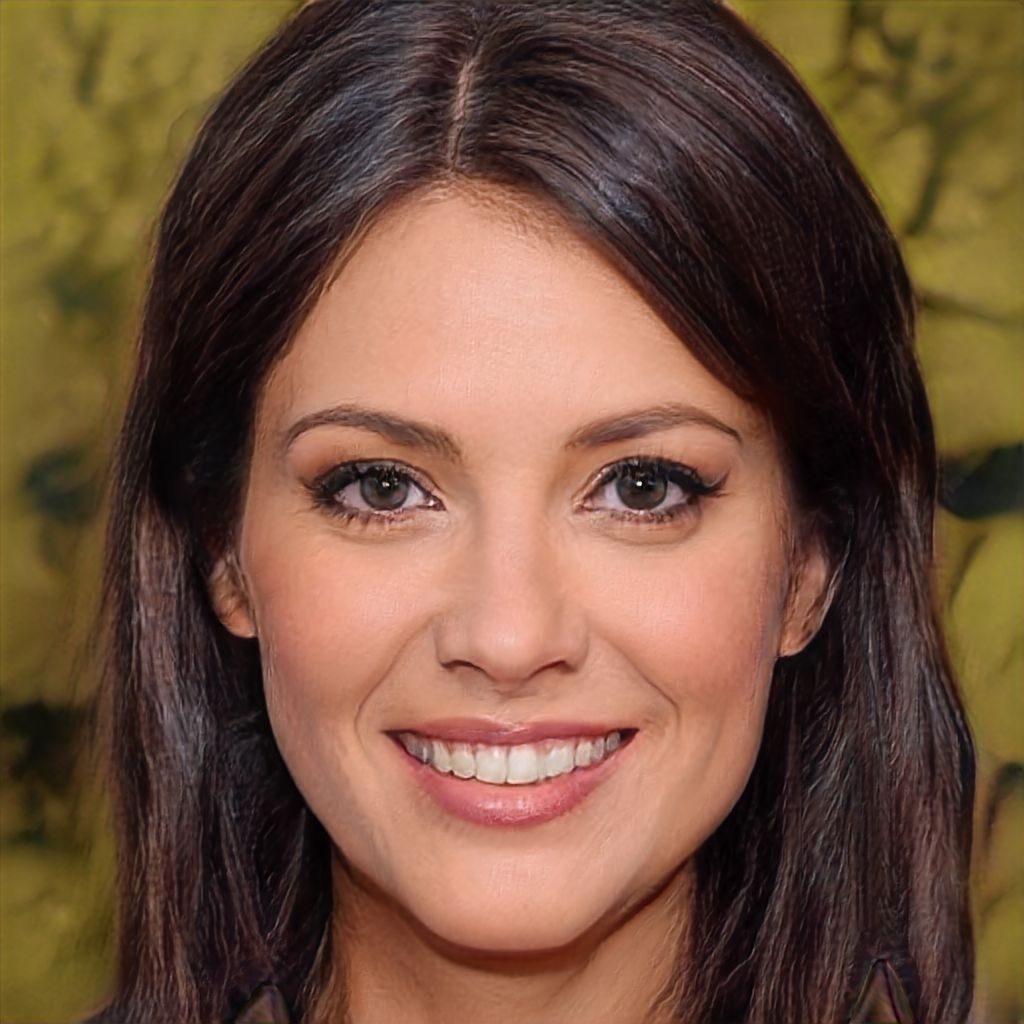}
        \caption{Generated with $\psi \approx 0$.}
    \end{subfigure}
    
    \begin{subfigure}[b]{0.5\textwidth}
        \centering
        \includegraphics[width=0.22\linewidth]{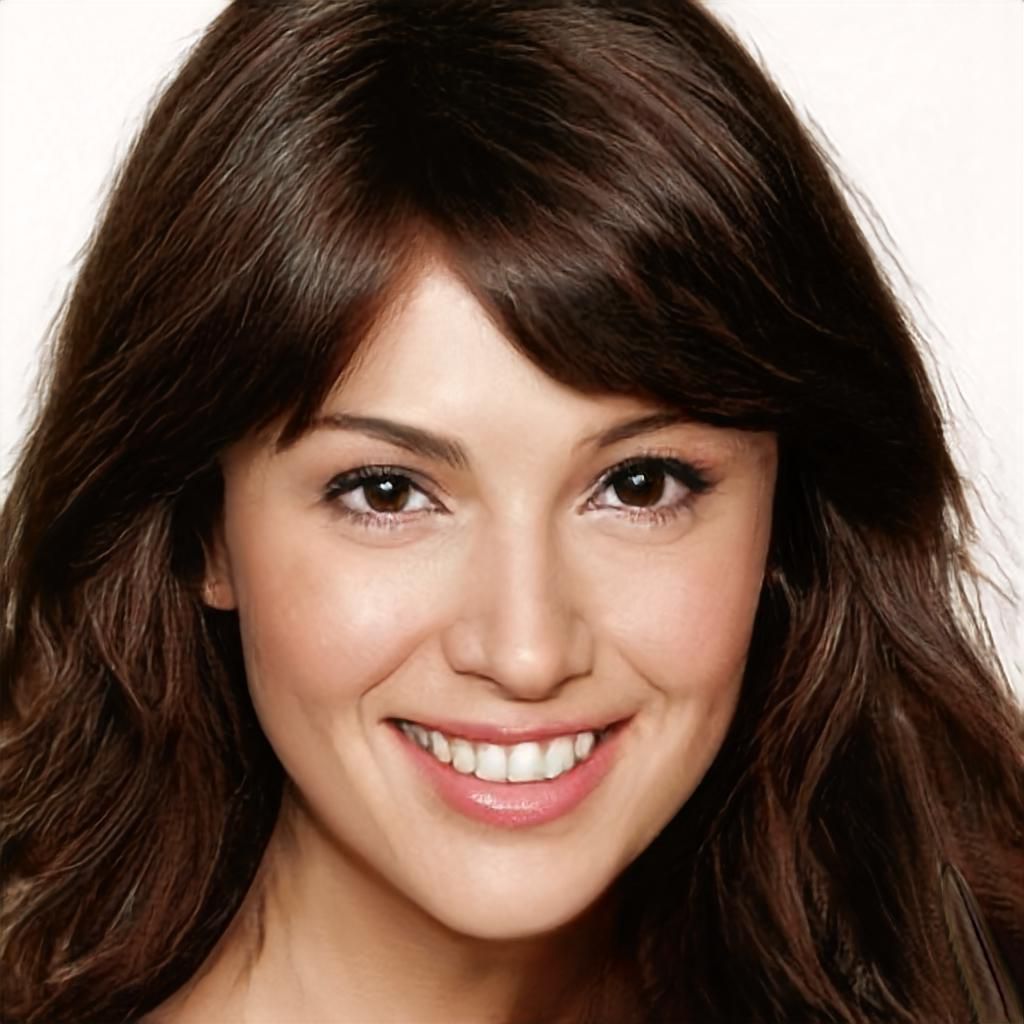}
        \includegraphics[width=0.22\linewidth]{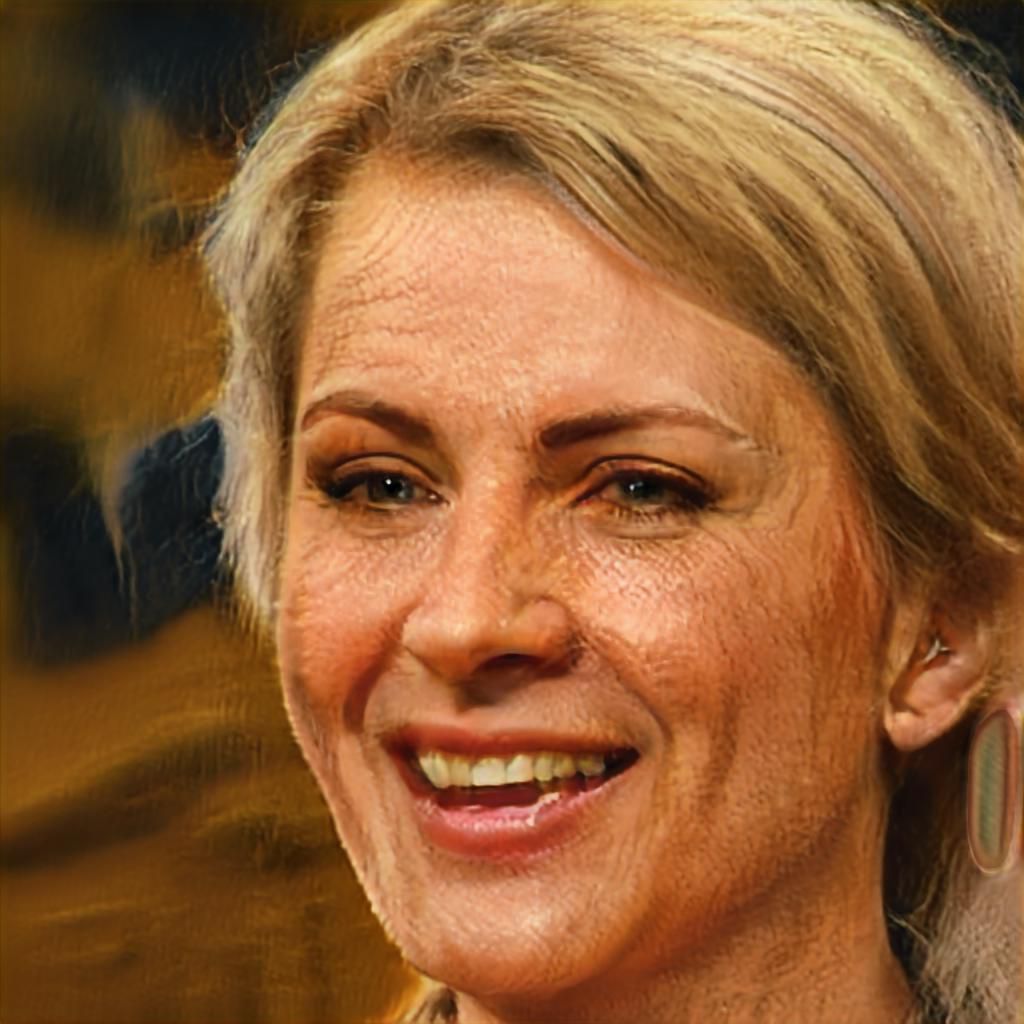}
        \includegraphics[width=0.22\linewidth]{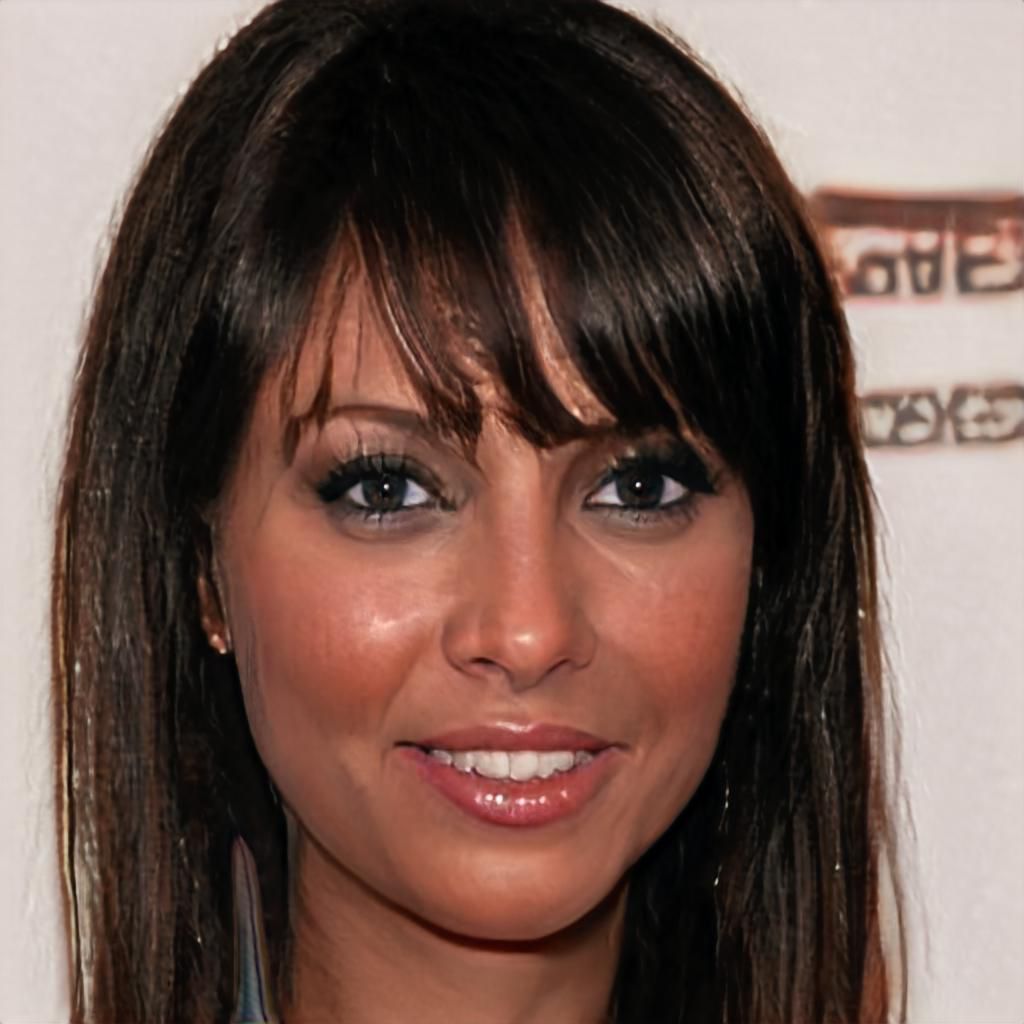}
        \includegraphics[width=0.22\linewidth]{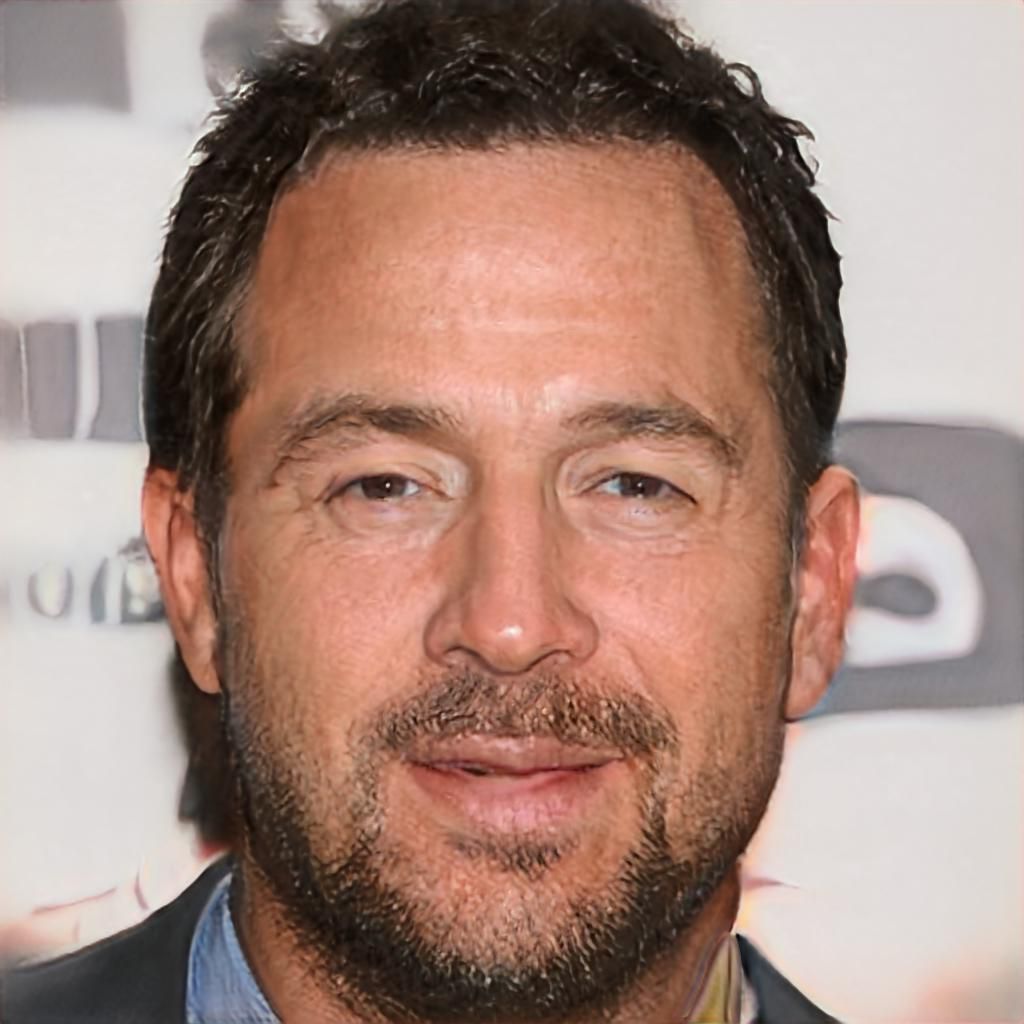}
        \caption{Generated with $\psi \approx 0.5$.}
    \end{subfigure}
    
    \begin{subfigure}[b]{0.5\textwidth}
        \centering
        \includegraphics[width=0.22\linewidth]{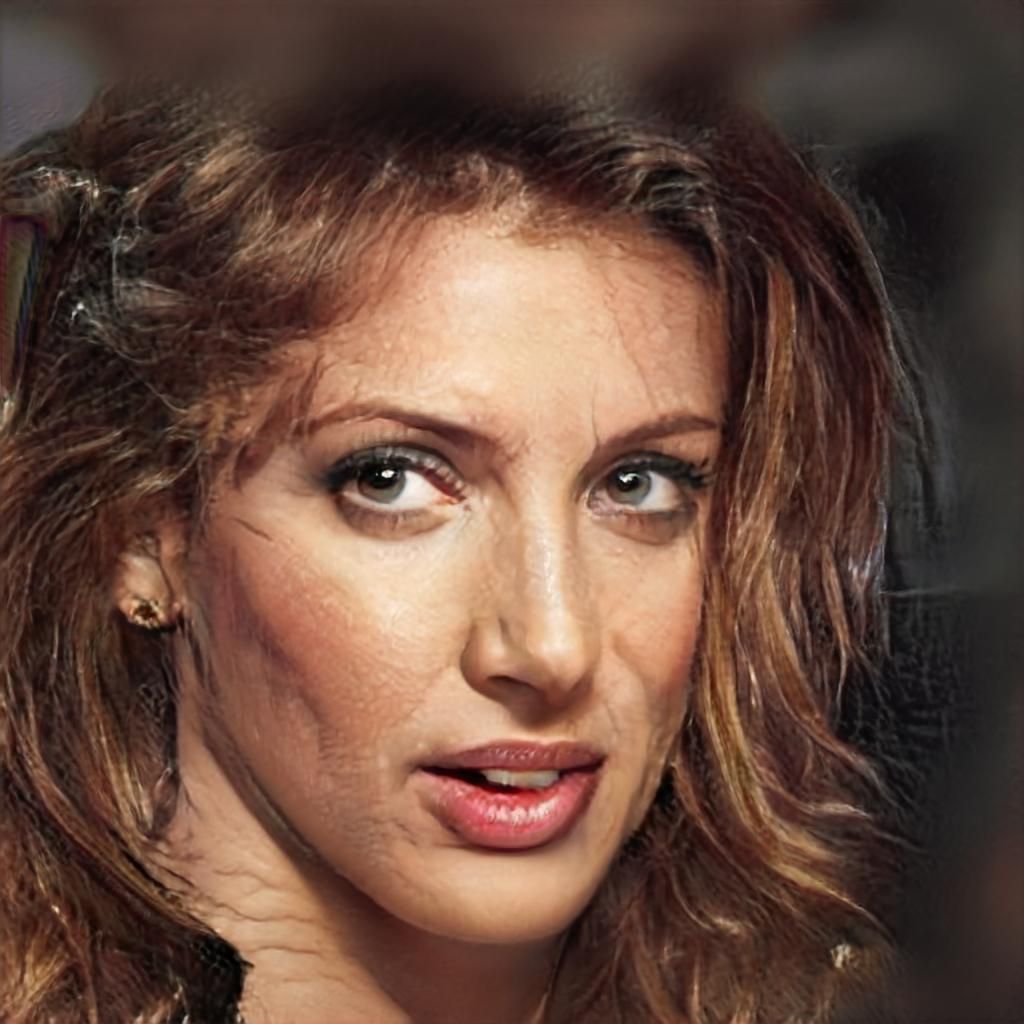}
        \includegraphics[width=0.22\linewidth]{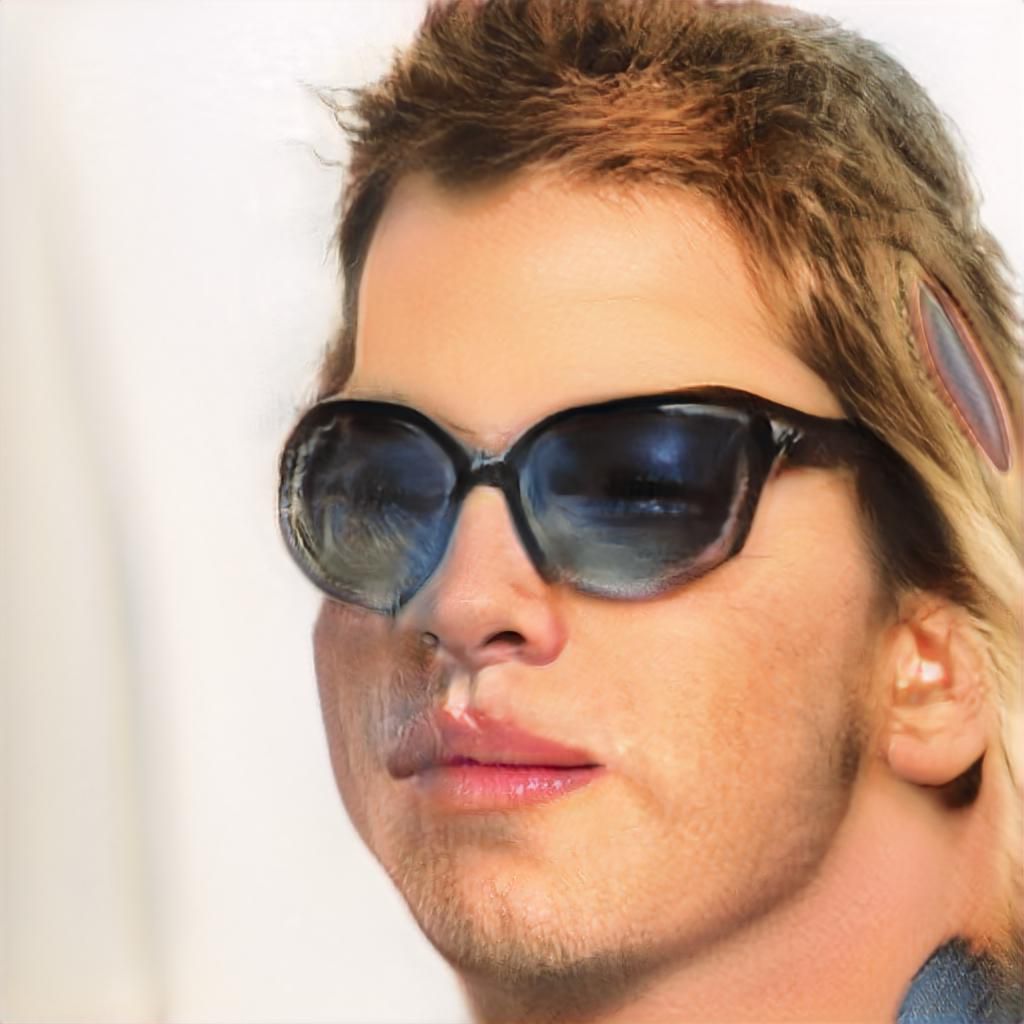}
        \includegraphics[width=0.22\linewidth]{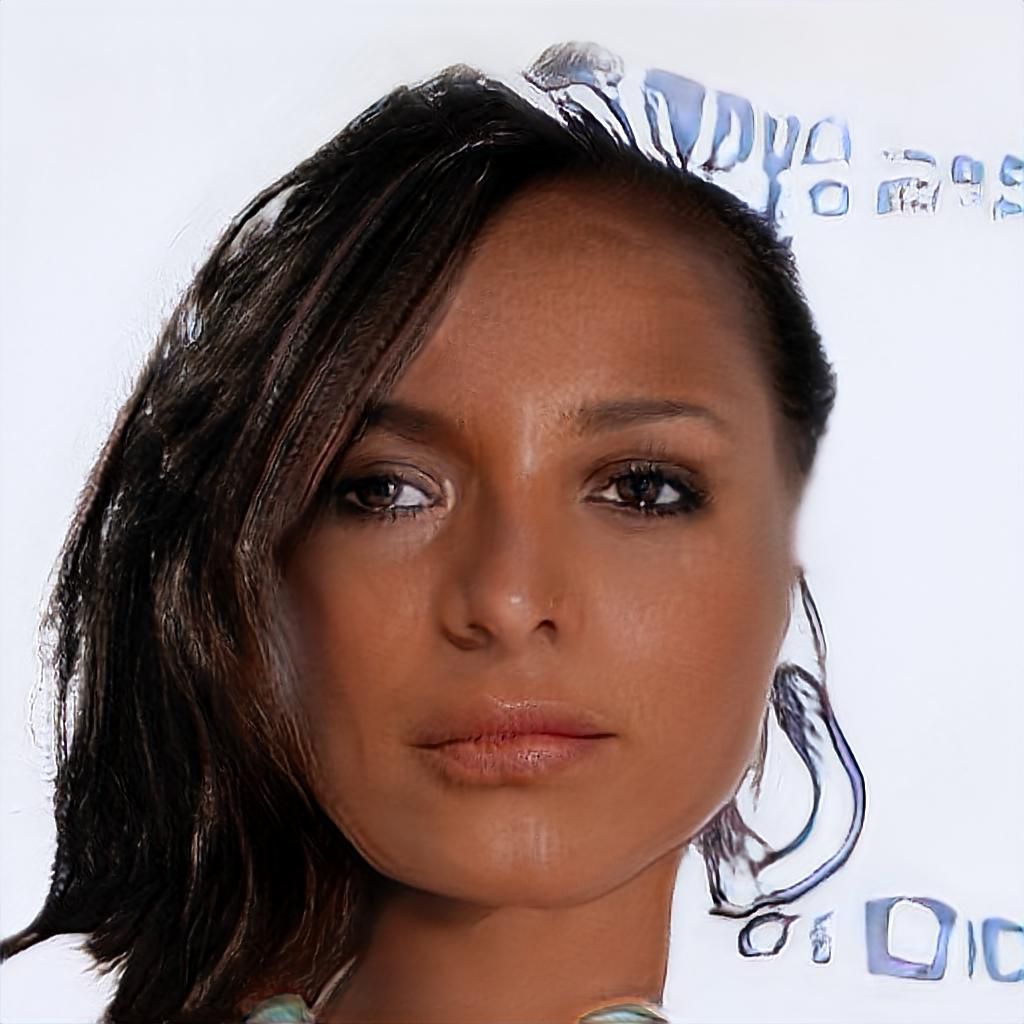}
        \includegraphics[width=0.22\linewidth]{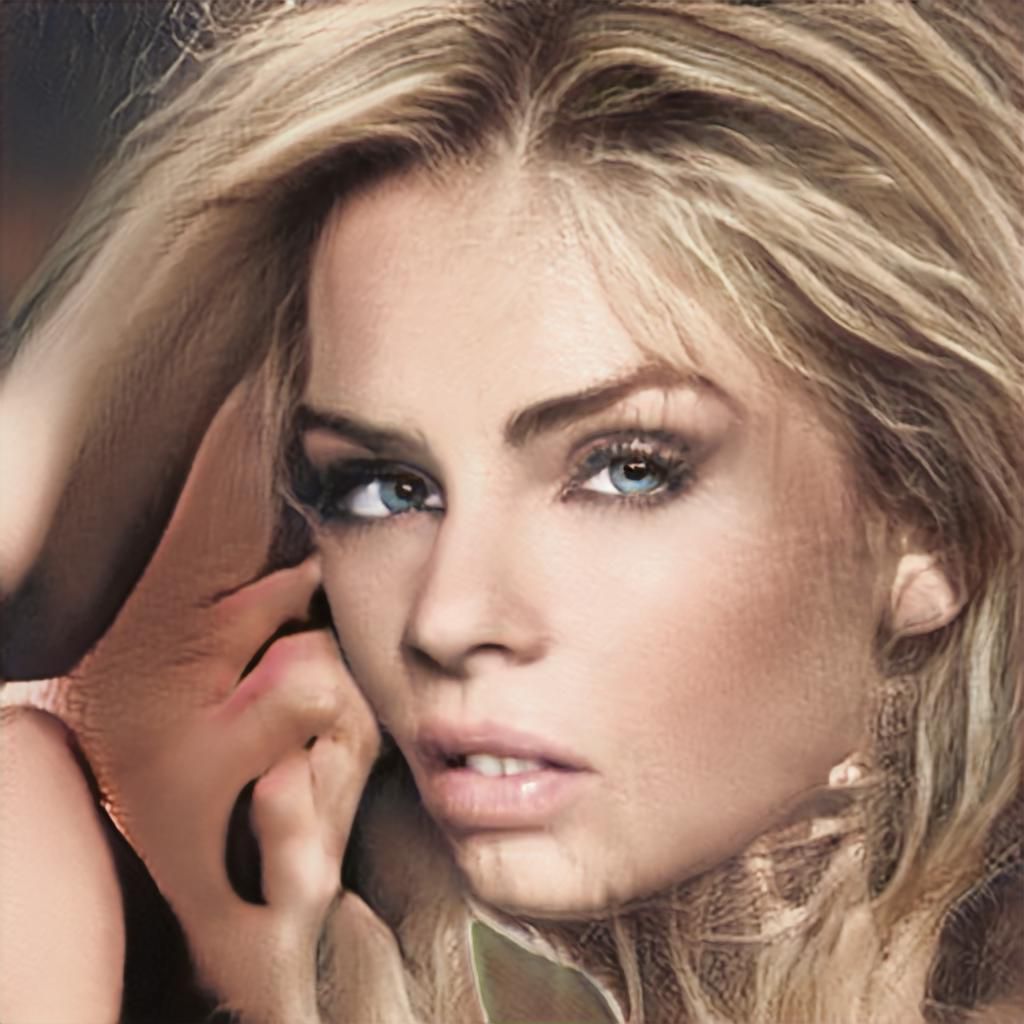}
        \caption{Generated with $\psi \approx 1$.}
    \end{subfigure}
    
    \caption{Manually selected images generated by StyleGAN\textsubscript{{\scriptsize \textit{CAHQ}}} with different quantities of the truncation trick. Note that this results in a trade-off between visually realistic (\ie with $\psi \approx 0$) and original/varied images (\ie with $\psi \approx 1$).}
    \label{fig:style_generated_examples}
\end{figure}

We generate images using a model pre-trained on CAHQ images, because there is no public dataset of such images. For generation we make use of the truncation trick [19], which refers to the \textit{stylistic} sampling radius (denoted by $\psi$) in the latent style vector. In other words, it refers to how much the style of the image to be generated should be similar to or divergent from the average style in the training data, where style refers to the characteristics of the full image, with a large focus on the person (\ie facial area) in the image, and a minor focus on the background. In our initial experiments, this latent sampling radius is uniformly sampled from $[0, 1]$. However, the set of images with $\psi \approx 0$ appears to be very homogeneous and predictable, without much geometrical variation. On the other hand, using a large value (\ie $\psi \approx 1$) results in original but unrealistic images with many artefacts. This is demonstrated in Figure~\ref{fig:style_generated_examples}. Both types of images do not represent real-world scenarios, where images are realistic and varied. Based on visual inspection of many images within the range of $\psi \in [0, 1]$, it seems that a good trade-off between quality and variety seems to be somewhere around $\psi \approx 0.5$. Thus, the dataset is generated using $\psi=0.5$, where each image is generated by passing a random noise vector (\ie no style transfer).

\subsection{Training procedure}
\label{subsec:training_procedure}
We train all models using the settings of [12], unless otherwise specified. We use a batch size of 64 for ForensicTransfer and 32 for Xception due to its higher memory demands. We evaluate two optimizers (SGD and Adam) and find that on average, SGD slightly outperforms Adam. Thus, we use SGD using a learning rate of 0.01, momentum of 0.9, and weight decay of 0.0001. We stop training after 3 epochs of no improvement, as we observe that overfitting tends to be slightly higher when we use 30 epochs as done by [12]. All models are trained on a single Nvidia Titan V GPU and take roughly 1-3 hours of training time per model. 

We evaluate the influence of a pre-trained Xception model on Imagenet in combination with pre-processing methods, and find that it performs worse with pre-trained weights. This is likely due to the large difference between images using for pre-training and our pre-processed images. Thus, we choose Xception to be trained from scratch, using weights randomly initialized from a normal distribution.

Lastly, we evaluate the influence of a random seed. Based on initial experiments, we observed some models and pre-processing methods to be unstable. For example, training with one random seed leads to a high test set accuracy, while another random seed leads to a much lower accuracy on the same dataset. This effect is even stronger for cross-model or cross-data test sets. To minimize the influence of a random seed on the results, 5 instances of each model\---pre-processing pair are trained, each initialized with another random seed. Then, the performance (\ie accuracies, not predictions) is averaged over the 5 instances. Every score reported in the results section is therefore an average of 5 model instances.

\section{Survey design}
\label{sec:survey_design}
This section describes six important elements of the survey design, including 1) the selection of images, 2) the gathering of participants, 3) the setup for testing the influence of feedback, 4) the setup for testing the influence of image resolution, 5) the image-questions, and 6) the meta-questions.

First, participants get to see an instruction screen with a motivation, the goal, and details of the survey, along with the guiding definitions of fake and real in this survey, as shown in \Cref{fig:real_fake_definitions}. These definitions are required since \textit{fake} is a vague definition and could also mean digitally edited (\ie photoshopped, or morphed together). Then, participants judge 18 images, and answer several meta-questions, as discussed later. Lastly, there is an overview page where participants see their total score (N out of 18 correct), and each of the 18 images, along with their own answer and the correct answers. Lastly, some information about the research is provided.

\begin{figure}[!htbp]
    \centering
    \noindent\fbox{%
    \parbox{0.97\columnwidth}{%
    \textbf{Real} image: taken with a camera, from a scene that really happened. Possibly post-processed, for example by adjusting colors. \\
    \textbf{Fake} image: a non-existing scene that is fully created by a computer. In other words, the person in the image does not exist.
    }%
    }
    \caption[Online survey definitions of real/fake.]{Provided definitions of real and fake images in the survey.}
    \label{fig:real_fake_definitions}
\end{figure}

\subsection{Selection of images}
\label{sec:c3_methods_s3_survey_design_ss2_dataset}
To achieve meaningful results, we use realistic and varied images. Therefore, we use real images from the FFHQ dataset, which is more varied and real-world than the CAHQ dataset. For fake images, we use the state-of-the-art StyleGAN\textsubscript{{\footnotesize \textit{\textbf{FFHQ}}}} images. Based on the findings of [51], along with our earlier experiments (\Cref{subsec:stylegan_cahq}), we select images generated using the truncation with $\psi=0.5$.

We \textit{manually} select 1000 good StyleGAN\textsubscript{{\footnotesize \textit{\textbf{FFHQ}}}} images and exclude images with very obvious artefacts such as large blobs, because these images would disturb the results. As shown by [20], these blob-like artefacts are already vanished in newer versions of StyleGAN, and including them would not give an accurate representation of how these images would be used in real-world scenarios (where images with obvious artefacts would be excluded). Note that in the selected survey, there are still smaller artefacts and other cues present that could be detected if one knows where to pay attention to.

Next, 1000 real images are randomly selected from the FFHQ dataset, of which a handful of images of celebrities is manually removed to avoid bias towards real, and a handful of images that look really weird or obviously photo-shopped is manually removed to avoid bias towards fake. Furthermore, this helps preventing potential situations where participants who do not fully understand the definition of fake (\eg thinking it means photoshopped) label a photoshopped image as fake. The resulting image pool consists of 1000 fake and 1000 real images, of which each participant sees 9 randomly selected images per class, in a random order.

\subsection{Participants}
\label{sec:c3_methods_s3_survey_design_ss3_participants}
In order to evaluate the detection capabilities of humans, a varied set of participants is tested. These participants vary in age, ethnicity, residence, education, AI-experience, \etc. They are approached through several mediums such as Facebook, Instagram, email, Reddit, and WhatsApp. The survey is conducted during May 2019, and results in 591 participants. Of these participants, 496 completed the whole survey, while 95 terminated early, which could be at any point in the survey. The participants who terminated early are excluded from all results. Participants who have not answered meta-questions are only excluded from results where that specific meta-question is relevant (\eg AI-experience). The amount of participants for different groups are shown in Table~\ref{tab:participant_group_sizes}. As shown, the distribution of AI-experience (little or much) within the control group and feedback group is roughly equal. 

\begin{table}[!htbp]
\small
\centering
\begin{threeparttable}
\tabcolsep=0.15cm
\begin{tabular}{l c c}
\toprule
Participant group                           & \multicolumn{2}{c}{Amount of participants} \\
\midrule
Started survey                              & 591 & - \\
\cmidrule{2-3}
Completed survey                            & 496 & 100.0\% \\
\cmidrule{2-3}
Control-group *                             & 263 & \phantom{0}53.0\% \\ % 53.02
Feedback-group *                            & 233 & \phantom{0}47.0\% \\ % 46.98
\cmidrule{2-3}
Filled in 'AI-experience'                   & 477 & \phantom{0}96.2\% \\ % 97.17
Little AI-experience \textdagger            & 218 & \phantom{0}45.7\% \\ % (34+91+93)=218
Much AI-experience \textdagger              & 259 & \phantom{0}54.3\% \\ %(136+123)=259
\cmidrule{2-3}
Control-group - Little AI-exp. \textdagger  & 117 & \phantom{0}24.5\% \\ % (21+47+49)=117
Control-group - Much AI-exp. \textdagger    & 136 & \phantom{0}28.5\% \\ % (76+60)=136
Feedback-group - Little AI-exp. \textdagger & 101 & \phantom{0}21.2\% \\ % (13+44+44)=101
Feedback-group - Much AI-exp. \textdagger   & 123 & \phantom{0}25.8\% \\ % (60+63)=123
\cmidrule{2-3}
Filled in 'image cues'                      & 481 & \phantom{0}97.0\% \\ % 96.98
\bottomrule
\end{tabular}
\caption[Overview of participant groups.]{Overview of participant amounts per group. * randomly assigned, thus not precisely balanced. \textdagger \ calculated as part of people who filled in 'AI-experience' (477).}
\label{tab:participant_group_sizes}
\end{threeparttable}
\end{table}

\subsection{Intermediate feedback}
\label{sec:c3_methods_s3_survey_design_ss4_feedback}
To evaluate whether participants are able to learn how to detect this type of fake images, two groups are constructed, to which respondents were randomly assigned. The first group is the control group and receives no intermediate feedback. Participants only get to see their results at the very end of the survey. The second group receives immediate feedback after labelling an image. This feedback is of the form \textit{Correct, the image was indeed [real/fake]} or \textit{Incorrect, the image was [real/fake]} and is shown above an image. Note that an image remains displayed in order to encourage people to see \textit{why} an image is real or fake, without giving specific instructions on how to recognize fake images.

\subsection{Image resolution}
\label{sec:c3_methods_s3_survey_design_ss5_resolution}
To evaluate whether image resolution influences the detection performance, three resolutions are evaluated: 256x256, 512x512, and 1024x1024 (the original size). They are resized using the standard interpolation method in web browsers. Each of these image sizes is tested with 3 real and 3 fake images, randomly chosen from the image pool, resulting in 18 images. Note that the random selection is without replacement, such that one participant cannot see the same image twice.

\subsection{Labelling images}
\label{sec:c3_methods_s3_survey_design_ss6_image_design}
\begin{figure}[!htbp]
    \centering
    \captionsetup{width=\linewidth}
    \includegraphics[width=0.47\textwidth]{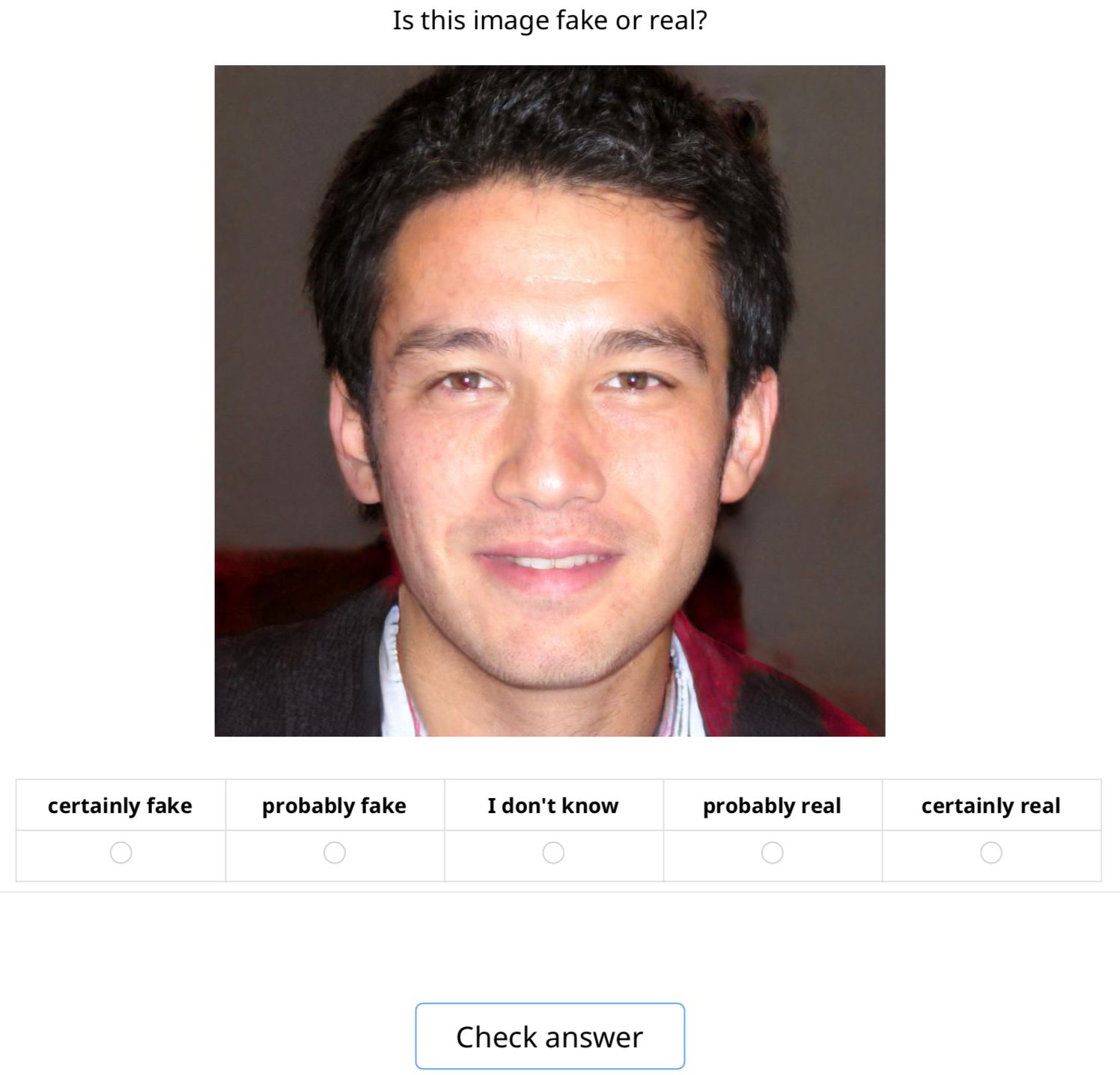}
    \caption{Screenshot taken from the online survey for a random fake image. Note that "Check answer" is only visible for participants from the feedback group, for the control group participants see a "Next" button. This image is generated by StyleGAN\textsubscript{{\footnotesize \textit{\textbf{FFHQ}}}}.}
    \label{fig:UI}
\end{figure}

Each participant sees 18 images sequentially and answers on a 5-point scale how certain it is that an image is real or fake. The answers are the following: \textit{certainly fake}, \textit{probably fake}, \textit{I don't know}, \textit{probably real}, \textit{certainly real}. Note that in the results, an answer is marked as correct if it is either the corresponding \textit{probably [real/fake]} or \textit{certainly [real/fake]} answer, and incorrect for the other three answers. A screenshot of our survey displayed in a web browser is shown in \Cref{fig:UI}.

There exists a website\footnote{\url{http://www.whichfaceisreal.com/}} where people can distinguish fake from real. On this website, a real and fake image are displayed next to each other, and users must select the one that is real. Such a setup is not appropriate for our survey, since we want to approximate real-world scenarios (\eg a social media timeline or forensic applications) where one would make a choice (consciously or unconsciously) based on \textit{one} image, and not a pair of images. Thus, we use an experimental setup with single images. 

\subsection{Meta-questions}
After labelling all images, participants have the choice to answer several meta-questions. Note that these questions are posed after the experiment itself to prevent any biases, and are not mandatory such that people can still finish the survey when they do not want to answer these questions. Most important are the questions about their AI-experience and cues they use to label images.

In order to evaluate the impact of domain knowledge, the amount of AI-experience is questioned using a 5-point scale, with the following answers: \textit{0 - none}, \textit{1 - heard of it}, \textit{2 - indirect experience}, \textit{3 - AI study}, and \textit{4 - AI-professional (PhD or work)}. We expect that this gives more meaningful results than having the answers \textit{little} and \textit{much}, since these answers might be too subjective for the participants. Based on their answers, we choose to group the first three into \textit{little} and the last two into \textit{much} AI-experience, where \textit{much} refers to AI-experts and \textit{little} refers to everyday people.

\begin{table}[!htbp]
\centering
\begin{threeparttable}
\begin{tabular}{l c c}
\toprule
Object cue              & Percentage \\
\midrule
Background              & 26.6 \\
Hair                    & 12.3 \\
Teeth                   & \phantom{0}8.7 \\
(A)symmetry             & \phantom{0}8.5 \\
Eyes                    & \phantom{0}7.7 \\
Composition             & \phantom{0}7.3 \\
Accessories / Context   & \phantom{0}7.3 \\
Ears                    & \phantom{0}6.1 \\
'Other'                 & \phantom{0}6.1 \\
Expression              & \phantom{0}5.4 \\
(Im)perfections         & \phantom{0}5.0 \\
Skin                    & \phantom{0}4.4 \\
Originality             & \phantom{0}2.4 \\
Mouth                   & \phantom{0}2.4 \\
\bottomrule
\end{tabular}
\caption{Object view image cues, ordered from most to least occurring. Percentage refers to how often the cue is mentioned as part of total amount of participants.}
\label{tab:object_cues}
\end{threeparttable}
\end{table}

In order to find out how humans can distinguish fake from real, respondents are posed the question: \textit{You have labeled 18 images on a scale from fake to real. What aspects in the images contributed to your decisions?} The choice for an open instead of closed question is simple: it is not desirable to bias the respondents towards certain answers. For example, if a list of \textit{eyes, nose, hair, \etc.} would be presented, they would easily reason further with that list in mind, resulting in, for example, a user input of \textit{mouth, teeth}. However, if the list would be too broad, such as \textit{eyes, nose, background, lighting conditions, \etc.}, the respondent might select multiple aspects without having actually thought of them during the experiment, resulting in biased backwards reasoning. The choice for an open question leads to a varied set of answers, as discussed in \Cref{sec:image_cues}.

\section{Image cues}
\label{sec:image_cues}
This final section discusses the image cues participants use to label an image as real or fake, based on their own answers after labelling all images. It becomes clear that the answers are very varied, ranging from specific answers such as \textit{blurry eyes} to more abstract answers such as \textit{something with the teeth} or \textit{unoriginal}.

\begin{table}[!htbp]
\centering
\begin{threeparttable}
\begin{tabular}{l c c}
\toprule
Display cue             & Percentage \\
\midrule
Blur                    & 40.1 \\
Artefacts               & 27.4 \\
Transitions             & 10.5 \\
Lighting / Shadow       & \phantom{0}9.3 \\
Reflections             & \phantom{0}4.8 \\
Details                 & \phantom{0}4.0 \\
Color                   & \phantom{0}2.4 \\
Focus / Depth of field  & \phantom{0}2.2 \\
'Other'                 & \phantom{0}1.6 \\
\bottomrule
\end{tabular}
\caption[Display view image cues.]{Display view image cues, ordered from most to least occurring. Percentage refers to how often the cue is mentioned as part of total amount of participants.}
\label{tab:display_cues}
\end{threeparttable}
\end{table}

Based on all answers, we decide to group them into two categories. First, there are \textit{object} cues, referring to \textit{physical} properties of the objects and background in the images. A few examples of such cues are \textit{weird shape of nose}, \textit{something with eye}, \textit{originality of background}, and \textit{expression}. The second category is referred to as \textit{display} cues, referring to \textit{how} these objects are displayed in an image as if they were captured by a camera. Several examples include \textit{artefacts}, \textit{blurry nose}, and \textit{lighting/shadows}. Clustering each of these cues is extremely difficult due to differences in jargon and specificity. Thus, our results should be taken with caution, since they approximate the distribution of image cues used by humans. Furthermore, some participants only answer with one example, while some answer with six examples, making this categorization even more difficult.

\begin{figure*}[!htbp]
    \centering
    \includegraphics[width=\textwidth]{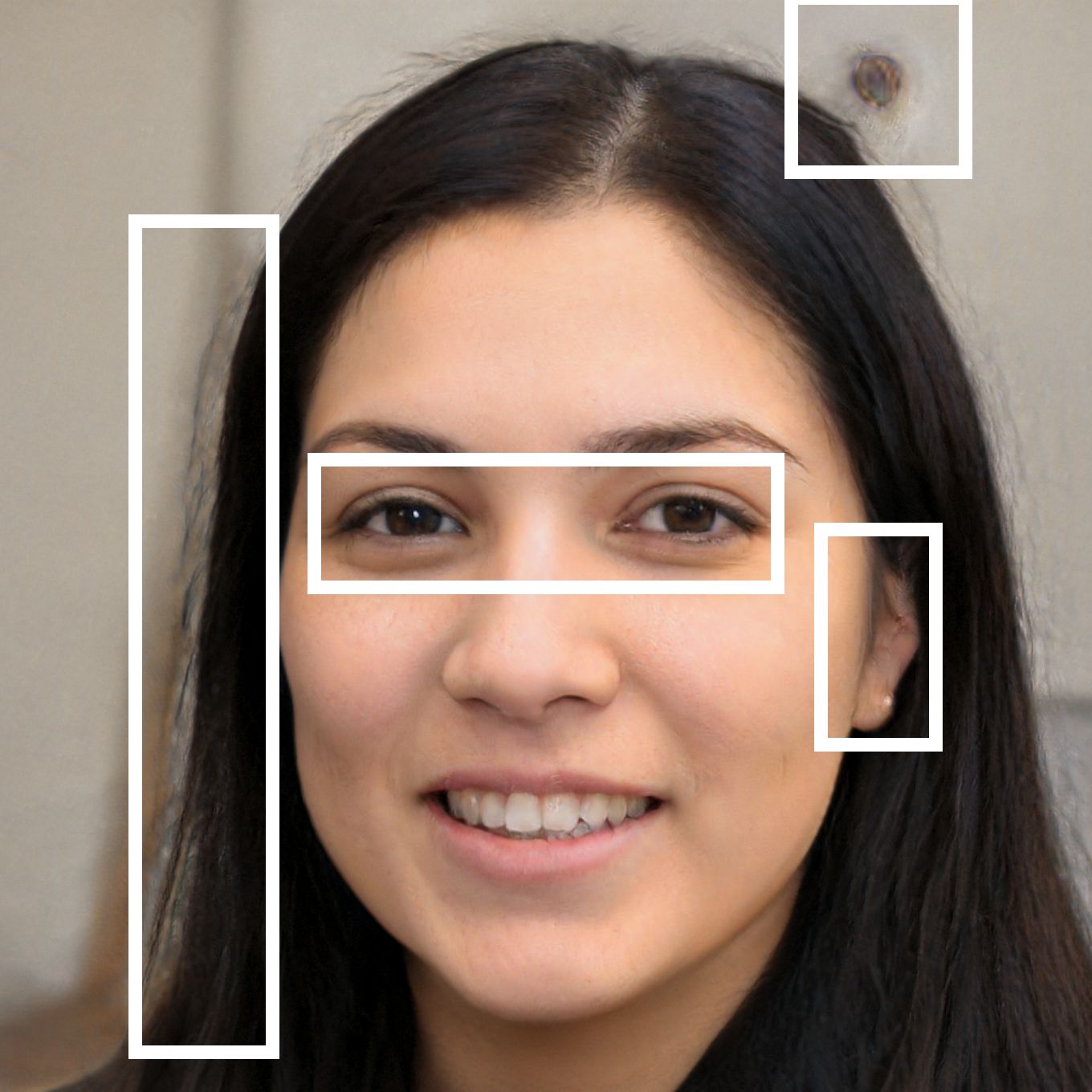}
    \caption{StyleGAN\textsubscript{{\footnotesize \textit{\textbf{FFHQ}}}} image with several unrealistic cues: 1) unnatural artefact (top-right), 2) blurry ear (right), 3) unrealistic/blurry hair (left), 4) asymmetric eyes (center). To elaborate on the last aspect: the iris colors, sizes, and shapes are slightly different between left and right eye. Furthermore, the pupil reflection only occurs at the left eye. When zooming in, artefacts (or lack of details) are better visible.}
    \label{fig:image_cues_examples}
\end{figure*}

The results of our clustering are shown in Table~\Cref{tab:object_cues} ('object cues') and Table~\Cref{tab:display_cues} ('display cues'). Lastly, we provide one visual example (\Cref{fig:image_cues_examples}) to refer to several of the cues shown in these tables.

%% file: ms.bbl
\begin{thebibliography}{10}\itemsep=-1pt

\bibitem{afchar2018mesonet}
Darius Afchar, Vincent Nozick, Junichi Yamagishi, and Isao Echizen.
\newblock Mesonet: a compact facial video forgery detection network.
\newblock In {\em WIFS}, 2018.

\bibitem{albright2019source}
Michael Albright and Scott McCloskey.
\newblock Source generator attribution via inversion.
\newblock In {\em CVPR Workshop on Media Forensics}, pages 96--103, 2019.

\bibitem{bayar2016deep}
Belhassen Bayar and Matthew~C Stamm.
\newblock A deep learning approach to universal image manipulation detection
  using a new convolutional layer.
\newblock In {\em ACM IHMS}, 2016.

\bibitem{chen2017image}
Bolin Chen, Haodong Li, and Weiqi Luo.
\newblock Image processing operations identification via convolutional neural
  network.
\newblock {\em arXiv preprint arXiv:1709.02908}, 2017.

\bibitem{chen2015median}
Jiansheng Chen, Xiangui Kang, Ye Liu, and Z~Jane Wang.
\newblock Median filtering forensics based on convolutional neural networks.
\newblock {\em IEEE Signal Processing Letters}, 22(11):1849--1853, 2015.

\bibitem{choi2018stargan}
Yunjey Choi, Minje Choi, Munyoung Kim, Jung-Woo Ha, Sunghun Kim, and Jaegul
  Choo.
\newblock Stargan: Unified generative adversarial networks for multi-domain
  image-to-image translation.
\newblock In {\em CVPR}, 2018.

\bibitem{chollet2017xception}
Fran{\c{c}}ois Chollet.
\newblock Xception: Deep learning with depthwise separable convolutions.
\newblock In {\em CVPR}, 2017.

\bibitem{cole2018blog}
Samantha Cole.
\newblock We are truly fucked: Everyone is making ai-generated fake porn now
  [blog post], 2018.
\newblock Accessed April 17, 2020.

\bibitem{cole2019blog_political}
Samantha Cole.
\newblock Deepfake of boris johnson wants to warn you about deepfakes [blog
  post], 2019.
\newblock Accessed April 17, 2020.

\bibitem{cole2019blog}
Samantha Cole, Emanuel Maiberg, and Jason Koebler.
\newblock This horrifying app undresses a photo of any woman with a single
  click [blog post], 2018.
\newblock Accessed April 17, 2020.

\bibitem{cozzolino2014image}
Davide Cozzolino, Diego Gragnaniello, and Luisa Verdoliva.
\newblock Image forgery detection through residual-based local descriptors and
  block-matching.
\newblock In {\em ICIP}, 2014.

\bibitem{cozzolino2017recasting}
Davide Cozzolino, Giovanni Poggi, and Luisa Verdoliva.
\newblock Recasting residual-based local descriptors as convolutional neural
  networks: an application to image forgery detection.
\newblock In {\em ACM IHMS}, 2017.

\bibitem{cozzolino2018forensictransfer}
Davide Cozzolino, Justus Thies, Andreas R{\"o}ssler, Christian Riess, Matthias
  Nie{\ss}ner, and Luisa Verdoliva.
\newblock Forensictransfer: Weakly-supervised domain adaptation for forgery
  detection.
\newblock {\em arXiv preprint arXiv:1812.02510}, 2018.

\bibitem{fridrich2012rich}
Jessica Fridrich and Jan Kodovsky.
\newblock Rich models for steganalysis of digital images.
\newblock {\em TIFS}, 7(3):868--882, 2012.

\bibitem{gilmer2019blog}
Marcus Gilmer.
\newblock As concern over deepfakes shifts to politics, detection software
  tries to keep up [blog post], 2019.
\newblock Accessed April 17, 2020.

\bibitem{goodfellow2014generative}
Ian Goodfellow, Jean Pouget-Abadie, Mehdi Mirza, Bing Xu, David Warde-Farley,
  Sherjil Ozair, Aaron Courville, and Yoshua Bengio.
\newblock Generative adversarial nets.
\newblock In {\em NeurIPS}, 2014.

\bibitem{iizuka2017globally}
Satoshi Iizuka, Edgar Simo-Serra, and Hiroshi Ishikawa.
\newblock Globally and locally consistent image completion.
\newblock {\em ACM Transactions on Graphics (ToG)}, 36(4):1--14, 2017.

\bibitem{isola2017image}
Phillip Isola, Jun-Yan Zhu, Tinghui Zhou, and Alexei~A Efros.
\newblock Image-to-image translation with conditional adversarial networks.
\newblock In {\em CVPR}, 2017.

\bibitem{jin2017cyclegan}
Xiaohan Jin, Ye Qi, and Shangxuan Wu.
\newblock Cyclegan face-off.
\newblock {\em arXiv preprint arXiv:1712.03451}, 2017.

\bibitem{karras2017progressive}
Tero Karras, Timo Aila, Samuli Laine, and Jaakko Lehtinen.
\newblock Progressive growing of gans for improved quality, stability, and
  variation.
\newblock In {\em ICLR}, 2018.

\bibitem{karras2019style}
Tero Karras, Samuli Laine, and Timo Aila.
\newblock A style-based generator architecture for generative adversarial
  networks.
\newblock In {\em CVPR}, 2019.

\bibitem{karras2019analyzing}
Tero Karras, Samuli Laine, Miika Aittala, Janne Hellsten, Jaakko Lehtinen, and
  Timo Aila.
\newblock Analyzing and improving the image quality of stylegan.
\newblock {\em arXiv preprint arXiv:1912.04958}, 2019.

\bibitem{kingma2018glow}
Durk~P Kingma and Prafulla Dhariwal.
\newblock Glow: Generative flow with invertible 1x1 convolutions.
\newblock In {\em NeurIPS}, 2018.

\bibitem{li2018detection}
Haodong Li, Bin Li, Shunquan Tan, and Jiwu Huang.
\newblock Detection of deep network generated images using disparities in color
  components.
\newblock {\em arXiv preprint arXiv:1808.07276}, 2018.

\bibitem{li2018identification}
Haodong Li, Weiqi Luo, Xiaoqing Qiu, and Jiwu Huang.
\newblock Identification of various image operations using residual-based
  features.
\newblock {\em IEEE Transactions on Circuits and Systems for Video Technology},
  28(1):31--45, 2018.

\bibitem{liu2018large}
Ziwei Liu, Ping Luo, Xiaogang Wang, and Xiaoou Tang.
\newblock Large-scale celebfaces attributes (celeba) dataset.
\newblock {\em Retrieved August}, 15:2018, 2018.

\bibitem{liu2020global}
Zhengzhe Liu, Xiaojuan Qi, Jiaya Jia, and Philip Torr.
\newblock Global texture enhancement for fake face detection in the wild.
\newblock In {\em CVPR}, 2020.

\bibitem{marra2018detection}
Francesco Marra, Diego Gragnaniello, Davide Cozzolino, and Luisa Verdoliva.
\newblock Detection of gan-generated fake images over social networks.
\newblock In {\em MIPR}, 2018.

\bibitem{marra2019gans}
Francesco Marra, Diego Gragnaniello, Luisa Verdoliva, and Giovanni Poggi.
\newblock Do gans leave artificial fingerprints?
\newblock In {\em MIPR}, 2019.

\bibitem{matern2019exploiting}
Falko Matern, Christian Riess, and Marc Stamminger.
\newblock Exploiting visual artifacts to expose deepfakes and face
  manipulations.
\newblock In {\em WACVW}, 2019.

\bibitem{mccloskey2018detecting}
Scott McCloskey and Michael Albright.
\newblock Detecting gan-generated imagery using color cues.
\newblock {\em arXiv preprint arXiv:1812.08247}, 2018.

\bibitem{mo2018fake}
Huaxiao Mo, Bolin Chen, and Weiqi Luo.
\newblock Fake faces identification via convolutional neural network.
\newblock In {\em ACM IHMS}, 2018.

\bibitem{nataraj2019detecting}
Lakshmanan Nataraj, Tajuddin~Manhar Mohammed, BS Manjunath, Shivkumar
  Chandrasekaran, Arjuna Flenner, Jawadul~H Bappy, and Amit~K Roy-Chowdhury.
\newblock Detecting gan generated fake images using co-occurrence matrices.
\newblock {\em Electronic Imaging}, 2019(5):532--1, 2019.

\bibitem{nightingale2017can}
Sophie~J Nightingale, Kimberley~A Wade, and Derrick~G Watson.
\newblock Can people identify original and manipulated photos of real-world
  scenes?
\newblock {\em Cognitive research: principles and implications}, 2(1):30, 2017.

\bibitem{park2019semantic}
Taesung Park, Ming-Yu Liu, Ting-Chun Wang, and Jun-Yan Zhu.
\newblock Semantic image synthesis with spatially-adaptive normalization.
\newblock In {\em CVPR}, 2019.

\bibitem{parkin2019blog}
Simon Parkin.
\newblock The rise of the deepfake and the threat to democracy [blog post],
  2019.
\newblock Accessed August 5, 2019.

\bibitem{pathak2016context}
Deepak Pathak, Philipp Krahenbuhl, Jeff Donahue, Trevor Darrell, and Alexei~A
  Efros.
\newblock Context encoders: Feature learning by inpainting.
\newblock In {\em CVPR}, 2016.

\bibitem{pevny2010steganalysis}
Tom{\'a}{\v{s}} Pevny, Patrick Bas, and Jessica Fridrich.
\newblock Steganalysis by subtractive pixel adjacency matrix.
\newblock {\em IEEE Transactions on information Forensics and Security},
  5(2):215--224, 2010.

\bibitem{porter2019blog}
Jon Porter.
\newblock Another convincing deepfake app goes viral prompting [blog post],
  2019.
\newblock Accessed April 17, 2020.

\bibitem{rahmouni2017distinguishing}
Nicolas Rahmouni, Vincent Nozick, Junichi Yamagishi, and Isao Echizen.
\newblock Distinguishing computer graphics from natural images using
  convolution neural networks.
\newblock In {\em WIFS}, 2017.

\bibitem{rao2016deep}
Yuan Rao and Jiangqun Ni.
\newblock A deep learning approach to detection of splicing and copy-move
  forgeries in images.
\newblock In {\em WIFS}, 2016.

\bibitem{rossler2019faceforensics++}
Andreas R{\"o}ssler, Davide Cozzolino, Luisa Verdoliva, Christian Riess, Justus
  Thies, and Matthias Nie{\ss}ner.
\newblock Faceforensics++: Learning to detect manipulated facial images.
\newblock In {\em ICCV}, 2019.

\bibitem{russakovsky2015imagenet}
Olga Russakovsky, Jia Deng, Hao Su, Jonathan Krause, Sanjeev Satheesh, Sean Ma,
  Zhiheng Huang, Andrej Karpathy, Aditya Khosla, Michael Bernstein, et~al.
\newblock Imagenet large scale visual recognition challenge.
\newblock {\em IJCV}, 115(3):211--252, 2015.

\bibitem{schetinger2017humans}
Victor Schetinger, Manuel~M Oliveira, Roberto da Silva, and Tiago~J Carvalho.
\newblock Humans are easily fooled by digital images.
\newblock {\em Computers \& Graphics}, 68:142--151, 2017.

\bibitem{sullivan2006steganalysis}
Kenneth Sullivan, Upamanyu Madhow, Shivkumar Chandrasekaran, and BS Manjunath.
\newblock Steganalysis for markov cover data with applications to images.
\newblock {\em IEEE Transactions on Information Forensics and Security},
  1(2):275--287, 2006.

\bibitem{sullivan2005steganalysis}
Kenneth Sullivan, Upamanyu Madhow, Shivkumar Chandrasekaran, and Bangalore~S
  Manjunath.
\newblock Steganalysis of spread spectrum data hiding exploiting cover memory.
\newblock In {\em Security, Steganography, and Watermarking of Multimedia
  Contents VII}, 2005.

\bibitem{szegedy2015going}
Christian Szegedy, Wei Liu, Yangqing Jia, Pierre Sermanet, Scott Reed, Dragomir
  Anguelov, Dumitru Erhan, Vincent Vanhoucke, and Andrew Rabinovich.
\newblock Going deeper with convolutions.
\newblock In {\em CVPR}, 2015.

\bibitem{wang2019cnn}
Sheng-Yu Wang, Oliver Wang, Richard Zhang, Andrew Owens, and Alexei~A Efros.
\newblock Cnn-generated images are surprisingly easy to spot... for now.
\newblock In {\em CVPR}, 2020.

\bibitem{yang2019exposing}
Xin Yang, Yuezun Li, Honggang Qi, and Siwei Lyu.
\newblock Exposing gan-synthesized faces using landmark locations.
\newblock In {\em ACM IHMS}, 2019.

\bibitem{yeh2017semantic}
Raymond~A Yeh, Chen Chen, Teck Yian~Lim, Alexander~G Schwing, Mark
  Hasegawa-Johnson, and Minh~N Do.
\newblock Semantic image inpainting with deep generative models.
\newblock In {\em CVPR}, 2017.

\bibitem{yu2018generative}
Jiahui Yu, Zhe Lin, Jimei Yang, Xiaohui Shen, Xin Lu, and Thomas~S Huang.
\newblock Generative image inpainting with contextual attention.
\newblock In {\em CVPR}, 2018.

\bibitem{yu2019attributing}
Ning Yu, Larry~S Davis, and Mario Fritz.
\newblock Attributing fake images to gans: Learning and analyzing gan
  fingerprints.
\newblock In {\em CVPR}, 2019.

\bibitem{zheng2019survey}
Lilei Zheng, Ying Zhang, and Vrizlynn~LL Thing.
\newblock A survey on image tampering and its detection in real-world photos.
\newblock {\em Journal of Visual Communication and Image Representation},
  58:380--399, 2019.

\bibitem{zhou2019hype}
Sharon Zhou, Mitchell Gordon, Ranjay Krishna, Austin Narcomey, Li~F Fei-Fei,
  and Michael Bernstein.
\newblock Hype: A benchmark for human eye perceptual evaluation of generative
  models.
\newblock In {\em NeurIPS}, 2019.

\bibitem{zhu2017unpaired}
Jun-Yan Zhu, Taesung Park, Phillip Isola, and Alexei~A Efros.
\newblock Unpaired image-to-image translation using cycle-consistent
  adversarial networks.
\newblock In {\em ICCV}, 2017.

\end{thebibliography}
